\documentclass[final]{article}

\usepackage{arxiv}

\usepackage[T2A,T1]{fontenc}
\usepackage[utf8]{inputenc}
\usepackage{CJKutf8}
\usepackage[russian, english]{babel}
\newcommand{\textcyr}[1]{{\fontencoding{T2A}\selectfont #1}}

\usepackage{mathptmx}

\usepackage{lmodern}

\usepackage{hyperref}       % hyperlinks
\usepackage{natbib}
\usepackage{url}            % simple URL typesetting
\usepackage{booktabs, arydshln}       % professional-quality tables
\usepackage{amsfonts}       % blackboard math symbols
\usepackage{nicefrac}       % compact symbols for 1/2, etc.
\usepackage{microtype}      % microtypography
\usepackage{amsmath}        % for \text and math environments
\usepackage{lipsum}		% Can be removed after putting your text content
\usepackage{graphicx}
\usepackage{doi}
\usepackage{float}
\usepackage{wrapfig}
\usepackage{multirow}
\usepackage{makecell}
\usepackage{fancyvrb}
\usepackage[most]{tcolorbox}

\usepackage{xcolor}
\definecolor{hallucination}{RGB}{220,0,0}
\definecolor{grammar}{RGB}{0,70,200}
\definecolor{logic}{RGB}{255,140,0}

\VerbatimFootnotes

\title{Gamayun’s Path to Multilingual Mastery: \\ Cost-Efficient Training of a 1.5B-Parameter LLM}

\author{ 
	\includegraphics[scale=0.35]{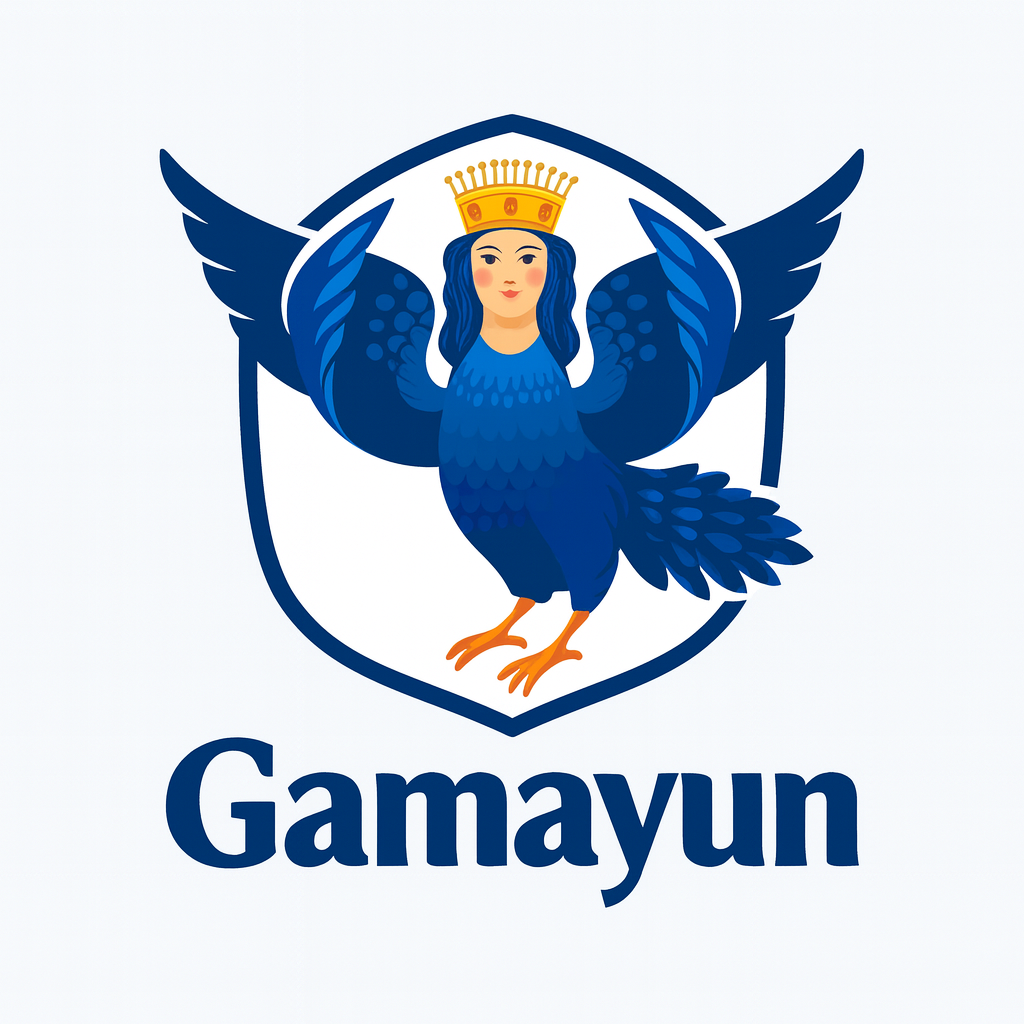}\\
    \hyperref[sec:contributors]{\hspace{2mm}Gamayun Team\hspace{2mm}}
}

\date{December 25, 2025}

\renewcommand{\shorttitle}{\textit{Gamayun} Technical report}

\hypersetup{
pdftitle={Gamayun-1.5B technical report},
pdfauthor={Gamayun Team},
pdfkeywords={Large Language Model, Efficient pre-training, Multi-lingual models, Russian LLM},
}

\begin{document}
\maketitle

\begin{abstract}
	We present Gamayun, a 1.5B-parameter multilingual language model trained entirely from scratch on 2.5T tokens. Designed for efficiency and deployment in resource-constrained environments, Gamayun addresses the lack of research on small non-English-centric LLMs by adopting a novel two-stage pre-training strategy: balanced multilingual training for cross-lingual alignment, followed by high-quality English enrichment to transfer performance gains across languages. Our model supports 12 languages, with special focus on Russian. Despite a significantly smaller training budget than comparable models, Gamayun outperforms LLaMA3.2-1B (9T tokens) on all considered benchmarks, and surpasses Qwen2.5-1.5B (18T tokens) on a wide range of English and multilingual tasks. It matches or exceeds Qwen3 (36T tokens) on most tasks outside advanced STEM, achieving state-of-the-art results in Russian, including the MERA benchmark, among the models of comparable size (1-2B parameters).
    
\end{abstract}

% keywords can be removed

\section{Introduction}

In this report, we present \textbf{Gamayun}, a multilingual language model with 1.5 billion parameters trained entirely from scratch.

While large language models (LLMs) continue to advance rapidly, there is growing demand for smaller models that can run in resource-constrained environments. Full pre-training from scratch is particularly valuable for organizations with large multilingual or domain-specific datasets, as it allows full control over the knowledge embedded in the model.

A key obstacle to this approach is the amount of training data required. Many recent methods rely on extremely large datasets, often generated by bigger models, or use distillation losses to mimic their outputs. While such techniques can be effective for specialized reasoning tasks like math and STEM, they significantly increase the cost of from-scratch training, making them less feasible in resource-constrained settings.

Compounding the problem, most existing small-scale LLMs are highly English-centric, with over 90\% of their training data in English. This leaves little research on how to design small multilingual models effectively, raising fundamental questions: How should languages be mixed? Should one language dominate while the others be “attached” to it? How do languages influence each other during training? 

To explore these questions and overcome the above challenges, we develop a competitive multilingual model under tight compute constraints. We limit the supported languages to 12, chosen according to business priorities and available data, to reduce multilingual interference. 
Special emphasis is placed on Russian, both in training and evaluation. The pre-training corpus is skewed toward English and Russian, reflecting resource availability and our primary evaluation targets, while still aiming for strong performance across all supported languages. In addition to language-specific sections of multilingual benchmarks, we evaluate on the MERA benchmark~\citep{fenogenova-etal-2024-mera}, developed by the Russian-speaking community to test Russian LLMs, and perform targeted validation of cultural knowledge.

Our initial experiments confirm the ``curse of multilinguality'': models trained on multiple languages underperform in their primary language compared to models trained solely on that language with the same primary-language data volume. This indicates that adding multilingual data reduces data efficiency for the main language. To address this, we propose a two-stage pre-training strategy, which improves both English and average multilingual performance without increasing the total amount of English-centric training. As a result, our model achieves state-of-the-art multilingual results among models of comparable size, despite using only a fraction of their pre-training budget.

As for general performance, we show that \textbf{2.5 trillion tokens} is sufficient to train a non-English-centric model under 2B parameters that performs well on general knowledge benchmarks, trailing top models only in advanced STEM tasks. Gamayun outperforms LLaMA3.2-1B (trained on 9T tokens) on all considered benchmarks and surpasses Qwen2.5-1.5B (trained on 18T tokens) on most English and multilingual tasks except MMLU. It is also comparable to or better than Qwen3 (trained on 36T tokens), except in MMLU and mathematical benchmarks, where Qwen3 likely benefits from heavy specialization.

\noindent\textbf{Main contributions:}
\begin{itemize}
    \item We demonstrate that 2.5T tokens are sufficient to train a multilingual model achieving state-of-the-art results on tasks, not requiring deep STEM knowledge, among models of comparable size.
    \item We propose a novel two-stage pre-training process: initial balanced multilingual training for cross-lingual alignment, followed by increased high-quality English data for performance transfer.
    \item We present \textbf{Gamayun 1.5B}, a compact LLM that matches strong competitors on English, achieves the best average performance on our multilingual evaluation, and is particularly strong in Russian, delivering state-of-the-art results on the MERA benchmark among models of comparable size.
    
\end{itemize}

\section{Related work}

\textbf{Small Large Language Models.}  
While early open-source model families typically did not include versions smaller than 7B parameters~\citep{touvron2023llama,touvron2023llama2,almazrouei2023falcon}, recent advances in architecture design~\citep{zhang2019root,su2024roformer,shazeer2020glu} and the availability of high-quality data~\citep{gao2020pile,weber2024redpajama,penedo2024fineweb} have significantly improved the performance of smaller models~\citep{subramanian2025small}. In addition to downsized versions of larger models, there are also independently developed small models designed specifically for edge device applications~\citep{abdin2024phi,tang2024pangu,team2024gemma}. The strong performance of these models on challenging benchmarks can be attributed to several factors: architectural innovations~\citep{tang2024pangu}, pretraining on large-scale synthetic data~\citep{abdin2024phi,qwen2.5}, post-training~\citep{meta_llama3_2_blog_2024} and distillation from larger models within the same family~\citep{team2024gemma}. For mathematical tasks, reinforcement learning with verifiable rewards has also shown promise when applied to smaller backbones~\citep{liu2025prorl,wang2504reinforcement,zeng2025simplerlzooinvestigatingtamingzero}. In our work, we focus on the careful selection of pre-training and post-training data, primarily utilizing existing open datasets, and introduce a novel multi-stage training scheme.
 
\textbf{Multilingual LMs.}
Modern LLM pretraining often relies heavily on a single dominant language, typically English, leading to severe imbalance to multi-lingual support~\citep{qin2025survey}. There have been deliberate efforts to build more balanced multilingual autoregressive models, such as XGLM~\citep{Lin2021FewshotLW}, which uses upsampling of low-resource languages to equalize representation, and mGPT~\citep{shliazhko2024mgpt}, a GPT-style model pretrained from scratch on dozens of languages to enable few-shot multilingual transfer. These early models explored the idea of allocating more resources to non-English languages and demonstrated the potential of multilingual LLMs (MLLMs) on cross-lingual benchmarks like XNLI~\citep{Conneau2018XNLIEC}, XCOPA~\citep{ponti2020xcopa}, and FLORES-101~\citep{Goyal2021TheFE}. However, they also revealed a trade-off: expanding to multiple languages often degraded performance on individual languages—an effect known as the "curse of multilinguality"~\citep{chang2023multilinguality}. More recent models like Qwen~\citep{qwen2.5}, LLaMA~\citep{touvron2023llama2}, and Gemma~\citep{team2024gemma} have shown stronger multilingual capabilities, sometimes even without explicit multilingual training. Compared to these newer models, XGLM and mGPT significantly underperform due to several factors, including lower data quality, smaller and less diverse training corpora, and outdated architecture and optimization strategies.

\textbf{Optimal pre-training compute.} 
Early studies relied on the Chinchilla scaling laws~\citep{hoffmann2022training}, which estimated the optimal trade-off between model size and training compute (FLOPs) for effective pretraining. However, modern small LLMs frequently exceed these guidelines, leveraging larger datasets and more compute than predicted~\citep{hu2024minicpm,abdin2024phi}. Moreover, previous work suggested that abilities like mathematical reasoning were emergent and unattainable for small models~\citep{wei2022emergent}. This assumption has since been challenged by recent progress in data curation, model distillation, and instruction tuning—techniques that allow smaller models to match or even outperform larger ones on reasoning and STEM benchmarks~\citep{qin2025survey}.

\textbf{LMs with focus on Russian language.} 
GigaChat~\citep{valentin2025gigachat} and YandexGPT~\citep{yandexgpt5_lite_2024}, the two prominent Russian-centric autoregressive LLM families, do not offer small variants below roughly 7B parameters, which may reflect the fact that Russian is relatively well-supported by multilingual models (often considered a mid-to-high-resource language)~\citep{qin2025survey}, so general multilingual LLMs already perform reasonably well on it. Consequently, strong Russian performance can be achieved more cheaply by adapting existing pretrained models instead of training from scratch, e.g., via tokenizer re-training and embedding propagation~\citep{tikhomirov2024facilitating} or through targeted fine-tuning to build bilingual instruction-following models~\citep{nikolich-etal-2024-vikhr}. However, adapted models by design lack deep, locally specific cultural knowledge~\citep{wu2025bitter}; in contrast, models pretrained directly on Russian data capture and preserve cultural context much more effectively. 

\section{Base model training}

\subsection{Gamayun 2T pre-training}
Given our resource constraints and focus on practical multilingual benchmarks, we trained the model on a curated dataset spanning 12 carefully selected languages: English, Russian, Arabic, Bulgarian, German, Spanish, French,  Italian, Polish, Portuguese, Thai and Chinese. This choice balances linguistic diversity and computational feasibility. It allowed us to cover multiple language families, scripts, and resource levels from low-resource languages like Thai to high-resource ones like Russian. Limiting the number of languages enabled deeper training per language and simplified data curation by focusing on languages with large, high-quality monolingual corpora.

\begin{figure}[b]
    \begin{minipage}{0.49\textwidth}
        \centering
        \includegraphics[width=0.9\textwidth]{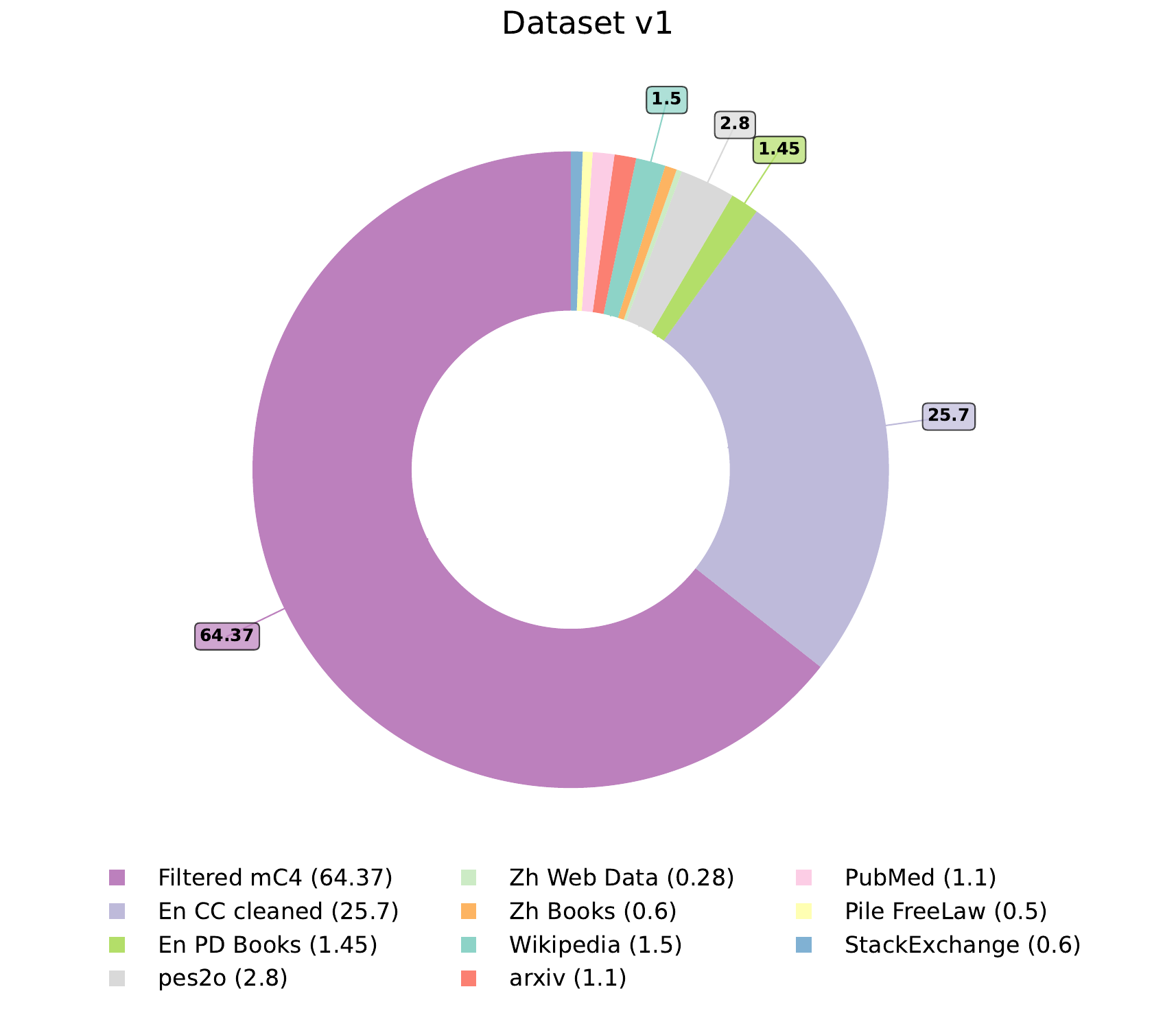}%{images/Dataset_v1_2.png}
    \end{minipage}
    \hfill
    \begin{minipage}{0.49\textwidth}
        \centering
        \includegraphics[width=0.9\textwidth]{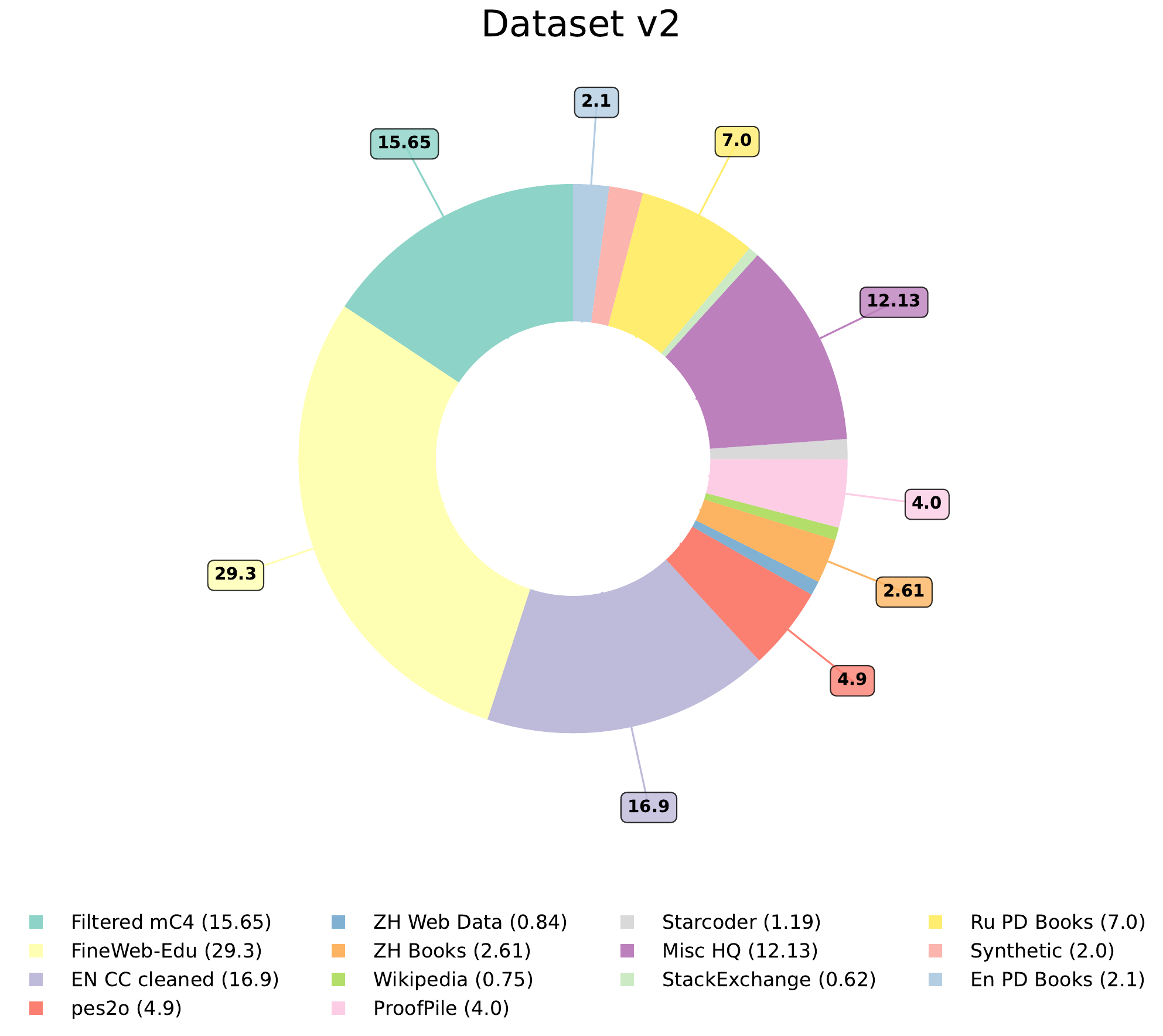}%{images/Dataset_v2_2.png}
    \end{minipage}
    \caption{\small
    First stage \emph{(left)} and second stage \emph{(right)}  data distributions.
    In the first stage, non-English data prevails. In contrast, the second stage becomes more English-centric; the amount of lower quality mC4 corpus decreased from 64\% to 16\%. %TODO: The breakdown of mC4 by languages is given in Appendix~\ref{app:DataComparison}.
    }\label{fig:data_stages}
\end{figure}

\textbf{Web-scale data.} To build the training corpus, we used mC4~\citep{xue-etal-2021-mt5} as a base due to its size. We applied additional filtering based on heuristics and quality scorers trained on the selected subsample of well-structured informative texts, and augmented it with high-quality open sources such as books (rich in structure and vocabulary), scientific papers (providing STEM knowledge), and code (from GitHub and StackOverflow for reasoning and planning) (see %Appendix~\ref{app:DataComparison} 
Fig.~\ref{fig:data_stages}). 

Filtering procedures include:
\begin{itemize}
    \item 5-gram model (trained on a Wikipedia subsample with kenlm library) perplexity cutoff;
    \item language-agnostic heuristics taken from~\citep{gopher};
    \item rule-based sub-document filtering of the metadata, found in various web texts, such as hashtags, links, timestamps, emails, malformed headers and footnotes etc.;
    \item finally, for Russian, we include a simplified pipeline based on \citet{fineweb}. Specifically, we used the model by \citet{tikhomirov2024facilitating}\footnote{\verb+https://huggingface.co/RefalMachine/ruadapt_qwen2.5_7B_ext_u48_instruct+} to collect a labeled data subsample ($\sim$ 300k documents), which was then used to train a classifier for large-scale application. The \verb+fasttext+ classifier demonstrated strong performance (F1 score $\sim$82) while remaining lightweight. We utilized an ensemble of two such classifiers, trained on different data subsamples and labeling prompts (see Appendix \ref{app:prompts}).
\end{itemize}

\begin{figure}[t]
   \centering
   \includegraphics[width=0.85\textwidth]{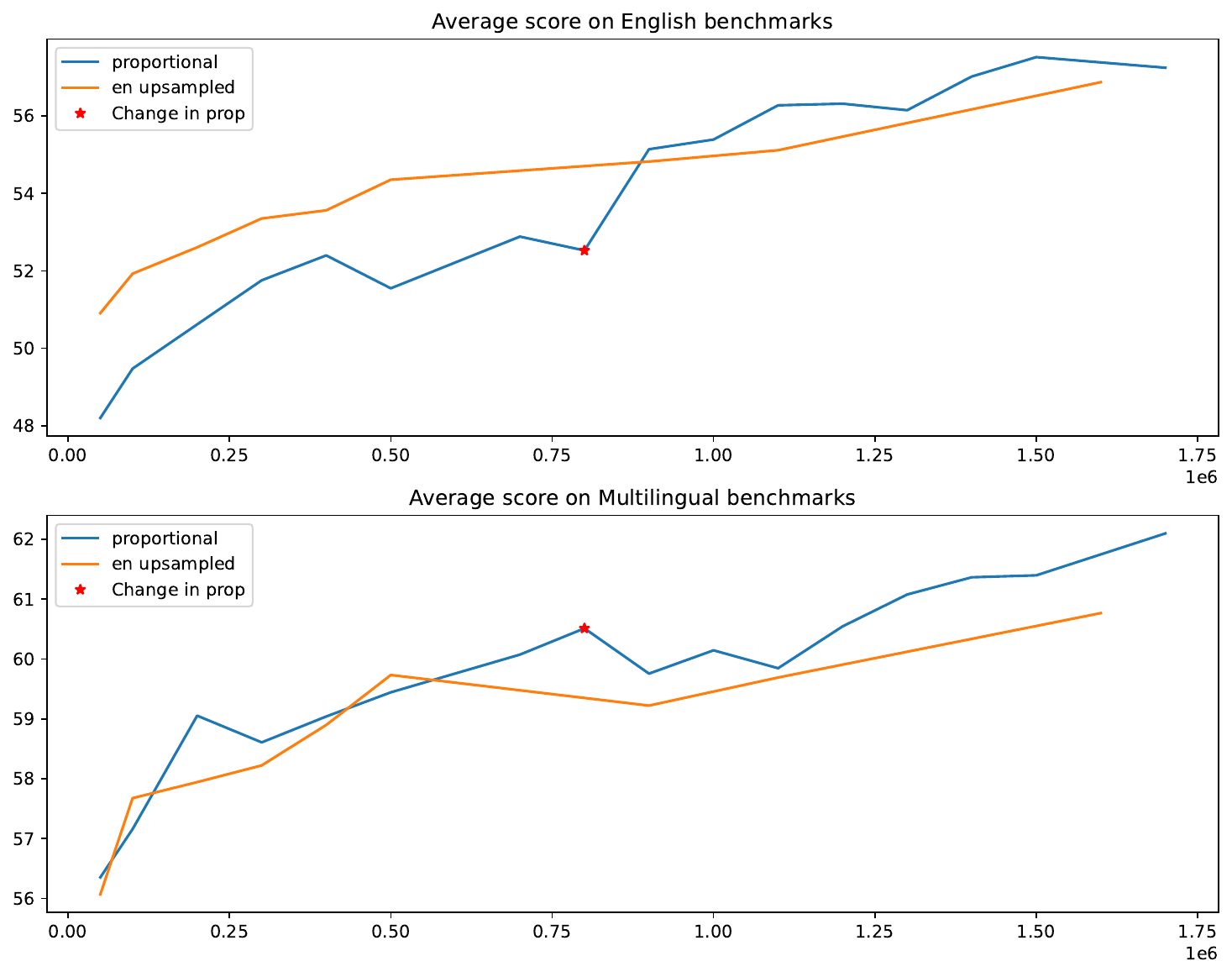}%{images/two_stages.png}
\caption{\small
Two-stage training of the 1.5B model: a balanced language mix in the first stage (to the left of the red point) ensures language alignment, which makes the increased proportion of high-quality English data in the second stage beneficial for all languages. We observe that the two-stage model (blue) outperforms the model trained without alignment stage (orange) in average accuracy on both English (top) and multilingual (bottom) benchmarks.}
\label{fig:two_stages_graph}
\end{figure}

\textbf{Language mixing.} Our experiments confirmed that naïve language mixing degrades performance on English benchmarks. To demonstrate it, we trained two 750M models on Wikipedia, one on English only, another on English plus three other languages. Despite identical English token exposure, the multilingual model had higher perplexity and worse LAMBADA~\citep{paperno-etal-2016-lambada} performance, illustrating the ``curse of multilinguality'' (see Appendix~\ref{app:pretraining} Fig.~\ref{fig:curse_of_mult}).

Then, we evaluated commonly used language sampling strategies designed to address language imbalance. Temperature scaling~\citep{conneau2020unsupervised} and UniMax sampling~\citep{chungunimax} improved low-resource languages but degraded high-resource ones like English and Russian, which were our focus (see Tab.~\ref{tab:sampling_ppl} in Appendix~\ref{app:pretraining}). Hence, we retained proportional sampling based on token counts.

Although mC4 data demonstrates reasonable pre-training quality, we found many issues in this dataset, especially for languages other than English. These issues included HTML artifacts, non-informative ads or stub web pages, adult content, etc. Further tests with mC4 showed that data quality varies significantly across languages, which influences training efficiency. This is demonstrated in Appendix~\ref{app:pretraining} Fig.~\ref{fig:mixing} using bilingual pre-training, comparing English-Russian and English-Chinese models trained with the same English subset. We observe that increasing the proportion of Russian data steadily improves performance on that language, while adding extra Chinese tokens beyond a certain threshold has minimal effect. More importantly, English loss degrades more in the latter case. This highlights that low-quality data not only underperforms but also reduces the effectiveness of high-quality data during training. Moreover, we observe that longer training on low-quality data doesn't give additional benefits.

We also found that training order highly affects outcomes. In~\citep{rukhovich2025commute}, there was demonstrated that mixed-language schedule can match single-language training only in the case when the languages are aligned in terms of commutativity of their loss flows. These theoretical findings supported our choice of two-stage training strategy: first, language alignment; then, quality-focused refinement.

\textbf{Multi-stage training with dynamic English proportion.}
Based on the above observations we split the training into two stages. We started training of Gamayun model with a dataset where English made up about 37\% of the tokens. At a later stage, we increased the English proportion to approximately 70\%. As a baseline, we trained a separate model with over 70\% English from the start, keeping the ratio fixed. In the first phase, as expected, the English-heavy baseline performed better on English benchmarks, while the more balanced model showed stronger results on multilingual tasks. However, after increasing the English share in the first model’s training, it not only caught up with the baseline on English benchmarks but also maintained its multilingual advantage (Fig.~\ref{fig:data_stages}). This experiment supports a dynamic data mixing strategy: start with a balanced multilingual distribution to align languages, then gradually shift toward high-quality and domain-rich data (e.g., STEM, code) in later stages. This approach allows high-quality English data to improve general performance across all languages.
% \subsection{FLOPS analysis}
% Here should be a compute calculation, in comparison to Chinchilla optimal numbers

\noindent\textbf{Key Takeaways}
\begin{itemize}
\item\textbf{Multilingual data quality gap:} Multilingual datasets exhibit a significant quality gap between English and other languages. This is likely caused by the poor performance of language-specific tools (such as crawlers, extractors, and filters), which makes training less effective for underrepresented languages.    
\item\textbf{Impact of low-quality data:} Low-quality data not only harms performance on the affected language but also reduces the effectiveness of high-quality data during training. This leads to worse results, even with the same total token count.    
\item\textbf{Two-stage training helps:} A two-stage approach, first using a representative multilingual mix, then focusing on high-quality data, effectively mitigates the ``curse of multilinguality'' and improves both generalization and performance on the target language.
\end{itemize}

\subsection{Annealing stage}
After the two-staged pre-training, we refine model's abilities during additional training on 400B tokens with increased ratio of high-quality data, such as STEM, code and synthetically generated data. Annealing on that data boosts model's performance on the downstream benchmarks, especially on those related to math. During annealing, we made several adjustments to the data mixture being used. We started with the pre-training mixture, then, increasing proportion of mathematical, general knowledge data. Empirically, we observe that addition of instruction data is also helpful. 

\textbf{Context length expansion.} During the second stage of annealing, we expand the context length from 2,048 tokens to 16,384 tokens using the ABF technique from \citep{xiong-etal-2024-effective}. The model's RoPE base frequency is increased from 10,000 to 500,000, and it is trained on additional 40B of tokens.

For this stage, we select long documents from the pre-training data covering web, books, code and STEM. We include 30\% sequences at maximum length and use shorter sequences for the rest part. This approach smoothly adapts model to an increased context length and preserves model's processing abilities across sequences of varying length.

\subsection{Model}
\subsubsection{Architecture}
For our model, we used LLaMA-like architecture, which employs SwiGLU activation, Rotary Positional Embeddings (RoPE), and RMSNorm applied before each transformer sub-layer (pre-normalization). 

Next, we chose a deep-and-narrow configuration, which has proven beneficial for smaller model sizes: 24 layers, 16 attention heads per layer, and a hidden size of 2048. This results in 1.5B parameters for the main model, plus 0.2B parameters from the tied input-output embeddings.

The final design choices of our model are summarized in Appendix~\ref{app:architecture}, Table~\ref{tab:architecture}.

\subsubsection{Tokenizer}
Another important design choice is the tokenizer. Key decisions include the training data, vocabulary size, and the algorithm (e.g., BPE, SentencePiece). Vocabulary size directly affects model size via the embedding layer. For instance, with a hidden size of 2048 and a 250K vocabulary, one embedding layer contains 500M parameters — over 30\% of a 1.5B model. Larger vocabularies also slow down training and inference due to the softmax bottleneck. However, they better support multiple languages and domain-specific terms. 

A crucial tokenizer property is \textit{fertility} --- the average number of tokens used per word. A tokenizer with higher fertility produces more tokens for the same sentence, which negatively affects computational efficiency (e.g., slower response times, higher memory usage), effective context length, semantic integrity, and linguistic fairness.

We evaluate fertility using the Universal Dependencies treebanks, a multilingual dataset with annotated word boundaries. In the Appendix, we compare the tokenizers of several modern LLMs to monolingual baselines: \texttt{ru} – ruBERT’s tokenizer, \texttt{en} – BERT’s tokenizer, \texttt{es} – GPT2-Spanish’s tokenizer, \texttt{de} – BERT-German’s tokenizer, and \texttt{zh} – BERT-Chinese’s tokenizer.

Monolingual tokenizers generally exhibit better fertility, but multilingual tokenizers with large vocabularies (e.g., XGLM) can achieve comparable results. We observe that LLaMA v3, GPT-4, and GPT-4o have fertility scores that are close to or even match those of monolingual tokenizers across most languages. Interestingly, later versions within the same model families (e.g., LLaMA, GPT) tend to increase vocabulary size and improve multilingual support compared to earlier versions.

Based on these considerations, we selected LLaMA 3's tokenizer, as it provides a good balance between vocabulary size and fertility.

\section{Post-training}
After training the Gamayun base model, we followed the current standard approach for maximizing performance and dialogue abilities via post-training through instruction tuning and preference learning. For post-training we use carefully selected existing datasets in addition to a new generated synthetic QA dataset. 

\subsection{Post-training data}
\subsubsection{Synthetic MCF Data}

Multiple-choice formulation (MCF), where the model is explicitly asked to select an answer from options 'A', 'B', 'C', or 'D', is a common structure in major LLM question-answering benchmarks such as MMLU~\citep{hendrycks2020measuring}. Small language models (<1.5--2B parameters) often struggle with MCF, although these difficulties can be mitigated by extended training. For example, the SmolLM2 report~\citep{allal2025smollm2smolgoesbig} observes that MMLU MCF accuracy begins to exceed the random baseline (25\%) only after 6T tokens of training. In contrast, the cloze-form version of the same benchmark (where the probability of the correct answer is evaluated directly) reaches its performance plateau significantly earlier. This suggests that the model's primary challenge lies in the task format itself, rather than in the underlying knowledge or reasoning capabilities.

Table~\ref{tab:MMLU_MCF} (Appendix~\ref{app:MMLU}) shows the MMLU MCF (5-shot) results for both base and instructed versions of various small models (1B--2B parameters). We observe that models with relatively limited pre-training can significantly improve their MCF performance during the instruction-tuning phase, resulting in a notable gap between the base and instructed versions (Gemma3~1B, LLaMA3.2~1B). In contrast, for models trained on large-scale corpora (>10T tokens), this gap is negligible or absent, indicating that their MCF ability is already well-developed during pre-training.

Since our pre-training budget was limited to 2.6T tokens, addressing MCF weaknesses during post-training became essential. 
%To mitigate this shortcoming, we added instruction datasets to the pretraining. Nevertheless, this approach proved its efficiency but it is insufficient.To encourage MCF capabilities further, 
We constructed a synthetic dataset of such tasks. For a subset of documents from FineWebEdu~\citep{penedo2024fineweb}, we generated four multiple-choice questions per document using Gemma3~12B~\citep{team2025gemma}. We introduced variation in both the answer indices (e.g., letters vs. numbers; uppercase vs. lowercase; different separators such as commas or parentheses) and the format of the answers (e.g., just the correct index vs. index followed by the full answer). This process resulted in approximately 380k QA pairs, comprising roughly 16\% of the full SFT dataset mixture.

Importantly, we generated QA pairs only from content already present in the pre-training data. This design choice aligns with the findings of~\citep{gekhman2024does}, which suggest that introducing novel data during the post-training phase can increase the model's susceptibility to hallucination.

\subsubsection{Instruction data}
To improve the instruction-following abilities of the Gamayun model, we used Flan-style data covering diverse task types. As shown in~\citep{mukherjee2023orca}, subsets of the Flan 2022 Collection~\citep{longpre2023flan} with GPT-4-generated answers can enhance model performance. We adopted the SlimOrca~\citep{SlimOrca} dataset, a curated 500k-example subset of OpenOrca~\citep{OpenOrca}, which aligns with the task distribution proposed in the Orca paper.

To strengthen reasoning in Russian, we also incorporated parts of the \texttt{ru\_instruct} dataset~\citep{ru-instruct-translated}, which includes several popular English instruction datasets translated into Russian.

\subsubsection{Math data}
A number of math datasets have recently been released, each targeting different types of problems. Based on insights from the SmolLM2 paper~\citep{allal2025smollm2smolgoesbig} and our experimental results, we selected NuminaMath-CoT~\citep{numina_math_datasets} for its strong impact on MATH~\citep{hendrycksmath2021}, and MetaMathQA~\citep{yu2023metamath} to improve performance on GSM8K~\citep{cobbe2021gsm8k}. Additionally, we included synthetic arithmetic problems~\citep{arithmetic} to help the model learn basic mathematical operations.

For obtaining better results in mathematics in Russian we added \texttt{kolmogorov-3} dataset~\citep{kolmogorov} which consists of carefully selected, checked and formatted olympiad-level russian math problems instructions. 

\subsubsection{Other Specialized Data}

For code generation, we used Self-OSS-Starcoder2-Instruct~\citep{wei2024selfcodealign} (50k high-quality Python instruction-response pairs from the SmolTalk dataset), 20k pairs from CodeAlpaca~\citep{codealpaca}, and 26.3k instructions from Russian-code~\citep{russiancode}. 

We also incorporated several other components from SmolTalk: Smol-Rewrite (rewriting), Smol-Constraints (detailed constraints similar to IFEval~\citep{zhou2023instruction}), and Smol-Summarization (summarization). Additionally, we included SystemChats2.0~\citep{systemchat2024} (system prompt support), Explore-Instruct-Rewriting~\citep{wan2023explore} (rewriting), and 100k instructions from OpenHermes2.5~\citep{OpenHermes2.5}, which has shown strong performance on knowledge benchmarks such as MMLU and MMLU-Pro.

%The final composition of our SFT dataset is provided in Table~\ref{tab:SFT_composition} (Appendix~\ref{app:post-training}). As shown in Table~\ref{tab:MixResults} (Appendix~\ref{app:DataComparison}), our post-training mixture achieves higher average performance compared to several widely used open instruction-tuning datasets.

\subsubsection{Alignment} 
Finally, we implemented the alignment training with preference optimization (see next section). For this stage, we start with \texttt{Llama-3.1--tulu-3-8b-preference-mixture}~\citep{Lambert2024TLU3P} dataset and include the on-policy data, i.e., text generations from SFT model that is used as a base model for preference finetuning. Namely, we generate 5\% responses to the queries of the dataset and use them as negative examples in DPO training.  

\subsection{Training Setup}

Our post-training process consists of supervised fine-tuning (SFT) and alignment phases. We summarize our hyperparameter settings in Table~\ref{tab:post-hyperparameters}.

\textbf{Supervised Fine-tuning (SFT).}
%To fine-tune the instructed Gamayun model, we used 8 Ascend 910B chips. We found that the results were the same as when using 8 NVIDIA V100 GPUs. 

The training used an effective batch size of 128, a maximum sequence length of 2048 tokens, and a learning rate of 1$\times$10\textsuperscript{-5}.

\begin{table}[h]
\caption{\centering Post-training hyperparameters for supervised fine-tuning (SFT) and alignment (DPO) phases.}
\centering
\begin{tabular}{@{}lll@{}@{}}
\toprule
\textbf{Hyperparameter} & \textbf{SFT} & \textbf{DPO}\\
\midrule
Learning Rate & 1 $\times$ 10\textsuperscript{-5} & 5 $\times$ 10\textsuperscript{-7}\\
Learning Rate Schedule & Linear & Linear \\
Batch Size (effective) & 128 & 64 \\
Max Token Length & 2048 & 1024  \\
Warm-up Ratio & 0.03 & 0.1 \\
Number of Epochs & 2  & 1 \\
DPO $\beta$ & -- & 5 \\ 
\bottomrule
\end{tabular}
\vspace{2mm}
\label{tab:post-hyperparameters}
\end{table}

\textbf{Alignment.}
For preference learning, we used Direct Preference Optimization (DPO)~\citep{rafailov2023direct}. After experimenting with several datasets and parameter settings, we mainly followed the approach proposed in~\citep{Lambert2024TLU3P}, which suggests using the length-normalized DPO variant from~\citep{meng2024simpo} (refer Table~\ref{tab:post-hyperparameters} for training hyperparameters).

The performance difference between the SFT and DPO versions is presented in Table~\ref{tab:DPOresults} (Appendix~\ref{app:DPOResults}).

\section{Evaluation}

\subsection{Benchmarks}
Because LLM evaluation procedures can include important details that affect comparability, we follow the standardized protocol from the LM Evaluation Harness initiative~\citep{eval-harness} for most datasets. We use zero-shot evaluation in all cases unless explicitly stated otherwise.
We selected a representative subset of tasks across all supported languages, with increased coverage for English and Russian.
For Gamayun model evaluation in English we choose the following common benchmarks: MMLU (5-shot)~\citep{hendryckstest2021} and MMLU-PRO (5-shot)~\citep{wang2024mmlu} for knowledge and STEM reasoning, ARC-easy~\citep{allenai:arc}, ARC-challenge~\citep{allenai:arc}, Ifeval~\citep{zhou2023instruction}, Hellaswag~\citep{zellers2019hellaswag}, Social-iqa~\citep{sap2019socialiqa}, and Winogrande~\citep{sakaguchi2021winogrande} for common-sense reasoning and general understanding. To measure mathematical ability we choose GSM8K~\citep{cobbe2021gsm8k}.

To test the performance in Russian language we use Russian subsets of GlobalMMLU~\citep{Singh2024GlobalMU} (which is the curated translation of MMLU), Belebele~\citep{bandarkar2023belebele}, XNLI~\citep{Conneau2018XNLIEC}, Xstorycloze~\citep{Lin2021FewshotLW}, Xwinograd~\citep{muennighoff2022crosslingual}, Multilingual ARC~\citep{dac2023okapi}, and Multilingual Hellaswag~\citep{dac2023okapi} as the representative set of general reasoning and language understanding benchmark. Besides, we use INCLUDE~\citep{Romanou2024INCLUDEEM}, which is a comprehensive knowledge- and reasoning-centric benchmark across 44 written languages aimed to evaluate multilingual LLMs for performance in the actual language environments where they would be deployed.

We also included the MERA benchmark~\citep{fenogenova-etal-2024-mera}, which is based on the Evaluation Harness methodology and is designed for evaluating Russian-centric models. The MERA score is calculated as the average performance across 15 tasks assessing different LLM abilities, using closed test subsets.

To specifically assess Russian cultural knowledge, we created RuBIN, an internal benchmark consisting of 5{,}000 multiple-choice questions on culturally specific topics, including proverbs, riddles, songs, and films. Unlike general-purpose benchmarks, RuBIN focuses exclusively on culturally relevant content (see Section~\ref{sec:culture} for details).

For other languages, we combined multilingual benchmarks (such as Belebele, INCLUDE, and MMLU-Global) with datasets specifically collected for a given language, such as ArabicMMLU, FrenchBench, and La Leaderboard (Spanish). See the complete list in Appendix~\ref{app:MBenchmarks}.

\subsection{Baseline models}
To make an appropriate comparison of Gamayun model we choose several state of the art models of the size between 1-2B parameters: Gemma3-1B~\citep{team2025gemma}; Llama3.2-1B~\citep{meta_llama3_2_blog_2024}; Olmo2-1B~\citep{olmo20242olmo2furious}; Qwen2.5-1.5B~\citep{qwen2024qwen25}; EuroLM-1.7B~\citep{martins2025eurollm} (which supports multiple European languages); SmolLM2~\citep{allal2025smollm2smolgoesbig}; Qwen3-1.7B~\citep{yang2025qwen3}.

\subsection{General results}
\begin{table}[b]
\caption{Results on English benchmarks.}
\centering
\setlength{\tabcolsep}{4.5pt}
\resizebox{\linewidth}{!}{
\begin{tabular}{l c c c c c c c c}
\toprule
Model & Gemma3 & Llama3.2 & Gamayun & Olmo2  & Qwen2.5 & EuroLM&  SmolLM2  & Qwen3  \\
\midrule
Size & 1B & 1.24B & 1.47B & 1.48B & 1.54B & 1.66B &  1.7B & 1.7B \\
Pre-training tokens & 2T & 9T & 2.6T & 4T & 18T & 4T & 11T & 36T \\
\midrule
\multicolumn{9}{c}{\textit{English}} \\
\midrule
MMLU (5-shot)    & 39.8 & 45.6 & 46.4  & 41.2 & \underline{59.7} & 28.3 &  49.2  & \textbf{60.2}  \\
MMLU-pro (5-shot)   &  15.2 & 19.6 & 24.9  & 16.5  &\underline{28.6} & 10.9 &   20.5 &  \textbf{32.1}  \\
Hellaswag  & 43.4   & 45.1 & 49.3   & \textbf{53.8} &  50.6 & 45.9  &   \underline{53.6}  & 46.2 \\
ARC-E     & 68.9 & 68.5 & \underline{73.7} & 70.9 & \textbf{74.1} & 71.3 &  68.8  & 72.6 \\
ARC-C     & 35.5 & 35.4 &  \underline{42.5}  & 40.8  & \textbf{43.3} & 35.9 &  38.7  & 40.1 \\
Ifeval & \underline{60.6}  & 52.7  & 56.2 & \textbf{69.1} & 44.0 & 23.7 & 51.4 & 48.9 \\ 
Social-iqa  &  41.8  & 41.7  &  48.2  & \underline{48.9}  & \textbf{50.6} & 44.8 &  44.8  & 45.2 \\
Winogrande  & 58.6  & 60.1 & 61.7  & \underline{63.9}  & 63.8 & 57.8 &  \textbf{68.5}  & 60.9 \\
\textbf{Average:} & 45.48 &	46.09 &	50.36 & 50.64 &	\textbf{51.84} &	39.83 &	49.44 & \underline{50.78} \\
\midrule
\multicolumn{9}{c}{\textit{Math}} \\
\midrule
%GSM8K (5-shot)   & 25.8 & 33.5  & 55.4 & \underline{62.3} & 34.2 & 15.6&  0.4 & \textbf{68.7} \\
%GSM8K-chat (5-shot)   & 31.3  & 38.2  & \underline{56.4} & \textbf{65.1} & 57.9 & 16.2 &  43.4  & 20.3\\
GSM8K (5-shot)   & 31.3 & 38.2  & 56.4 & \underline{65.1} & 57.9 & 16.2 &  43.4 & \textbf{68.7} \\
\bottomrule
\end{tabular}
}
\label{tab:main_results}
%\vspace{0.5em}
\end{table}

As shown in Table~\ref{tab:main_results}, Gamayun achieves competitive average performance, outperforming LLaMA3.2, EuroLM, and SmolLM2 across most benchmarks despite a significantly smaller pre-training budget. While the latest Qwen iterations (2.5 and 3), leveraging 18T and 36T tokens respectively, demonstrate exceptional MMLU scores, this advantage diminishes on aggregate averages. For instance, the 13.8-point MMLU gap between Qwen 3 and Gamayun narrows to just 0.42 points, suggesting a potential distributional skew in the Qwen pre-training corpus toward MMLU-aligned tasks.
Mathematical evaluation on GSM8K categorizes models into three tiers: ``math expert'' (Qwen3 and Olmo2, 65--69\%), ``strong performers'' (Qwen2.5 and Gamayun, 56--58\%), and less capable models (<45\%). 

Across all benchmarks, these results indicate that raw pre-training volume does not strictly scale with performance; notably, Gamayun outperforms SmolLM2 (11T) and LLaMA3.2 (9T) in most tasks. In terms of the cost-performance ratio, Olmo2 (4T) is the primary competitor; however, as Olmo2 and SmolLM2 are English-centric, they are excluded from further multilingual evaluation.

Multilingual results in Table~\ref{tab:MResults} (Appendix~\ref{app:MResults}) demonstrate that our model achieves the highest average scores or a second best result on most of the languages. Remarkably, this performance is attained without multilingual post-training (beyond English and Russian), validating our pre-training language alignment strategy. Furthermore, the model shows a substantial lead in translation tasks with non-English targets (see Appendix~\ref{app:MResults}), highlighting robust cross-lingual generalization acquired during the pre-training phase.

\begin{table}[t]\label{tab:MResults}
\centering
\setlength{\tabcolsep}{4.5pt}
\caption{Multilingual results.}
\begin{tabular}{l c c c c c c c c}
\toprule
Model & Gemma3 & Llama3.2 & Gamayun  & Qwen2.5 & EuroLM & Qwen3  \\
Size & 1B & 1.24B & 1.47B & 1.54B & 1.7B & 1.7B \\
\midrule
\multicolumn{7}{c}{\textit{Russian}} \\
\midrule
Arc\_ru & 28.1 & 24.4 & \textbf{34.9} & \underline{29.2} & 28.6 & 28.2 \\ 
Belebele-ru & 43.0  & 49.7 & \underline{70.4} & \textbf{71.3} & 29.4  & 67.9 \\
Hellaswag\_ru & 35.8 & 32.2 & \textbf{42.4} & 37.0 & \underline{37.1} & 35.6\\
INCLUDE-ru & 36.4 & 37.0 & \underline{46.4} & \textbf{52.9} & 23.4 & \textbf{52.9}\\ 
MMLU-ru (5-shot) & 34.5  & 35.9  & 41.2 & \underline{47.5} & 27.1 & \textbf{50.3}  \\
Xnli-ru  & 43.3 & 44.0 & \textbf{47.8} & 45.1 & \underline{46.0} & 40.6 \\
Xstorycloze &  62.8  & 55.8  & \textbf{68.3} & 63.1& \underline{63.7} & 59.1\\
Xwinograd  & 61.3 & 57.8 &  \textbf{66.7} & \textbf{66.7} & 61.0  & \underline{62.2} \\
MERA & 30.0 & 29.8  & \textbf{38.1} & 35.7 &  23.2 & \underline{36.4}  \\
\midrule
\multicolumn{7}{c}{\textit{Average performance by language}} \\
\midrule
Russian & 41.7 & 40.7 & \textbf{50.7} & \underline{49.8} & 37.7 & 48.1 \\
Arabic & 45.3 & 41.8 & \textbf{57.2} & \underline{53.7} & 39.2 & 49.6 \\
Bulgarian & 33.6 & 32.1 & \textbf{48.4} & 41.7 & 31.6 & \underline{45.7} \\
German & 40.4 & 43.9 & \underline{55.2} & \textbf{55.7} & 32.6 & 49.9 \\
Spanish & 49.3 & 46.2 & \textbf{55.7} & \underline{55.0} & 48.6 & 53.2 \\
French & 48.5 & 46.9 & \underline{55.2} & \textbf{55.5} & 47.5 & 52.6 \\
Italian Evalita-LLM & 36.3 & 35.7 & \underline{41.3} &  \textbf{43.3} & 34.0 & 40.2 \\
Italian ItaBench trans-mc & 53.1 & 50.7 & \textbf{54.2} & \underline{53.4} & 53.3 & 53.3 \\
Polish Open Leaderboard &  \underline{46.8}  & 34.3  &  \textbf{53.2} &  40.3 &  33.3 & 43.2  \\
Portuguese & 53.7 & 37.0 & \underline{61.1} & \textbf{64.6} & 34.1 & 54.9 \\
Thai & 40.2 & 41.2 & 47.3 & \textbf{54.5} & -- &  \underline{50.6} \\
Chinese & 44.3 & 49.6 & 55.6 & \textbf{68.7} & -- & \underline{63.4} \\
\bottomrule
\end{tabular}
%\vspace{0.5em}
\end{table}

\subsection{Russian-focused evaluation}\label{sec:russian}

Although standard benchmarks are designed to encompass various facets of LLM knowledge, they may not fully reflect a model's suitability for practical, everyday tasks, particularly in multilingual contexts. In this section, we perform a deeper evaluation of Russian language capabilities: first, by measuring performance on MERA~\citep{fenogenova-etal-2024-mera} benchmark specifically designed for Russian LLMs; second, by evaluating the model's culture-specific knowledge; and finally, by assessing the quality of long-form generation in Russian.

\subsubsection{MERA results} 
For comparison, we select models in the 1--2B parameter range from the MERA leaderboard. Notably, there are currently no other publicly available models of comparable size that were pre-trained with an explicit Russian focus; consequently, most competitive baselines in this regime are Russian-oriented adaptations of multilingual LLMs, and we follow this comparison setup. 
The selected models include: \textit{ruadapt}~\citep{tikhomirov2024facilitating}, derived from Qwen 2.5 through tokenizer re-training and fine-tuning; \textit{QVikhr-3-1.7B}~\citep{nikolich2024vikhr} and \textit{meno-tiny-0.1}~\citep{bondarenko2024meno}, which are fine-tuned versions of Qwen 3; and \textit{VikhrLlama}~\citep{nikolich2024vikhr}, based on Llama 3.2. We note that \textit{QVikhr-3-1.7B} and \textit{VikhrLlama} utilize the same post-training methodology. For a complete perspective, we also include the respective multilingual base models (Qwen 2.5, Qwen 3, and Llama 3.2) in our comparison.

As demonstrated in Table~\ref{tab:mera_metrics}, Gamayun outperforms all other models on the MERA benchmark. In general, adapted models demonstrate limited gains over their base versions; only the methodology employed by the Vikhr team yields a marginal improvement. Detailed performance metrics across all 15 tasks are provided in Table~\ref{tab:MERAResults} (Appendix~\ref{app:MERAResults}).

\begin{table}[t]
\caption{Russian-focused evaluation for selected models.}
\centering
\setlength{\tabcolsep}{5pt}
\begin{tabular}{lcccccccc}
\toprule
Metric & \shortstack{Gamayun\\1.5B} & \shortstack{Qwen2.5\\1.5B} & \shortstack{ruadapt\\(Qwen2.5)} & \shortstack{Qwen3\\1.7B} & \shortstack{QVikhr\\(Qwen3)} & \shortstack{meno-tiny-0.1\\(Qwen3)} & \shortstack{Llama3.2\\1.24B} & \shortstack{VikhrLlama\\(Llama3.2)} \\
\midrule
MERA & \textbf{38.1}  & 35.7  & 33.7  & 36.4  & \underline{36.8}  & 35.2  & 29.8  & 30.0  \\
%- no math & \textbf{36.3}  & \underline{35.7}  & 34.8  & 33.1  & 32.9  & 30.4  & 25.9  & 30.6  \\
RuBIN & \textbf{35.2}  & 32.0  & 31.0  & 32.0  & \underline{33.0}  & 29.0  & 25.0  & 25.0  \\
\bottomrule
\end{tabular}
\label{tab:mera_metrics}
\end{table}

\subsubsection{Russian culture-specific evaluation}\label{sec:culture}
Language data imbalance can limit coverage of target-language culture in pre-training. Such deficits rarely surface in global knowledge benchmarks, which often overlook local history, literature, and pop culture. INCLUDE attempts to fill this gap, but its Russian subset contains much non-cultural content (e.g., driving exam excerpts, medical protocols).

We therefore developed RuBIN, a multiple-choice benchmark in the format “How does this string end?”, covering memes, songs, proverbs, literary quotations, and popular films. Questions are designed to be answerable by native speakers without external resources. We did not add this data to pre-training, assuming it appears naturally in Russian web sources.

In Table~\ref{tab:mera_metrics}, Gamayun achieves the highest RuBIN score, which we attribute to a higher proportion of Russian web data in pre-training. Several models fail to exceed random-choice accuracy; adaptation often provides little benefit and may even reduce performance (e.g., meno vs.\ Qwen3).  
These findings suggest that multilingual pre-training from scratch is essential for small models to excel in culturally grounded reasoning.

\subsubsection{Long-form generation.}
Examples of Gamayun's answers, in comparison to Qwen2.5, are presented in Appendix~\ref{sec:examples}. Both models can provide correct short answers (e.g., Examples~\hyperref[ex4:planets]{4}, \hyperref[ex5:jupyter]{5}), but tend to hallucinate in longer explanations (Example~\hyperref[ex1:hamlet]{1}). Both also solved the simple mathematical task, though Qwen’s output was formatted more clearly (Example~\hyperref[ex3:math]{3}).

Overall, small LMs can successfully handle tasks requiring exact and concise output, but are not reliable sources of general-form information. 

Compared to Gamayun, Qwen more frequently struggles with complex Russian morphology, producing non-existent or inappropriate word forms (e.g., Example~\hyperref[ex2:anna]{2}). This underscores the effectiveness of our Russian-focused training.

\section{Conclusion}

We present Gamayun, a model trained with modest computational resources yet delivering strong multilingual performance, surpassing several strong competitors.
Our two-phase pre-training starts with a balanced language mix for cross-lingual alignment, then re-mixes the corpus to increase the share of high-quality, knowledge- and reasoning-rich data from high-resource languages. Experiments confirm that this strategy effectively transfers skills and knowledge from high-resource to lower-resource languages.

Training relied primarily on open-source datasets, complemented by numerous small-scale experiments to identify an optimal mix. We believe these findings can inform the development of other small, domain- or language-specific models.
We avoided large-scale distillation from bigger models, except for a small fraction of publicly available synthetic data (e.g., Cosmopedia). In cost-sensitive settings, producing large synthetic datasets can be prohibitive, as distillation typically requires inference from teacher models 10–20$\times$ larger, multiplying costs. It is also impractical when working with proprietary datasets unseen by any large LLM.

Our results show that strong general language and reasoning abilities can be learned on a relatively small budget. However, Gamayun still lags behind models trained with 10$\times$ more computation in complex tasks such as advanced knowledge and mathematics. Closing this gap is a priority for future work.

While Gamayun is designed to be broadly capable across many languages, we place a deliberate emphasis on Russian: we invest extra effort in Russian data collection and cleaning, and conduct a deeper Russian-centric evaluation beyond standard multilingual benchmarks. This focus yields practical insights into real-world language competence, including more fluent handling of complex Russian morphology compared to strong multilingual competitors, and stronger culture-specific knowledge than Russian post-trained adaptations. Overall, our results suggest that, although post-training multilingual models for a target language is appealing, it may be insufficient in practice: fine-tuning does not necessarily improve cultural understanding or reasoning over culture-specific topics.

\section{Limitations and ethical considerations}
Our evaluation focuses on benchmarks, not free-form generation. While Gamayun produces fluent and well-structured answers, it remains prone to hallucination, a common limitation of small LLMs. We caution against using its long-form outputs without verification and are committed to reducing hallucinations in future releases.

\section*{Contributors}\label{sec:contributors}

Alexander Podolskiy, 
Semen Molokov, 
Timofey Gerasin, 
Maksim Titov,
Alexey Rukhovich,
Artem Khrapov,
Kirill Morozov,
Evgeny Tetin,
Constantine Korikov,
Pavel Efimov,
Polina Lazukova, 
Yuliya Skripkar,
Nikita Okhotnikov,
Irina Piontkovskaya,

Meng Xiaojun,
Zou Xueyi,
Zhang Zhenhe.

\section*{Acknowledgements}\label{sec:contributors_ack}

We are grateful for useful comments and interesting data-related research ideas to
Elena Entina,
Laida Kushnareva,
Kristian Kuznetsov,
Alexander Kharitonov.

\bibliographystyle{abbrvnat}
\bibliography{references}

\clearpage
\appendix

\section{On the Name \textit{Gamayun}}

\textit{Gamayun} (\textcyr{Гамаюн})\footnote{\url{https://en.wikipedia.org/wiki/Gamayun}} is a prophetic bird of Russian folklore. It is a symbol of wisdom and knowledge and lives on an island in the mythical east, close to paradise. She is said to spread divine messages and prophecies, as she knows everything of all creation, gods, heroes, and man. 

\section{Multi-lingual pretraining}\label{app:pretraining}

\begin{figure}
     \centering
     \includegraphics[width=0.7\textwidth]{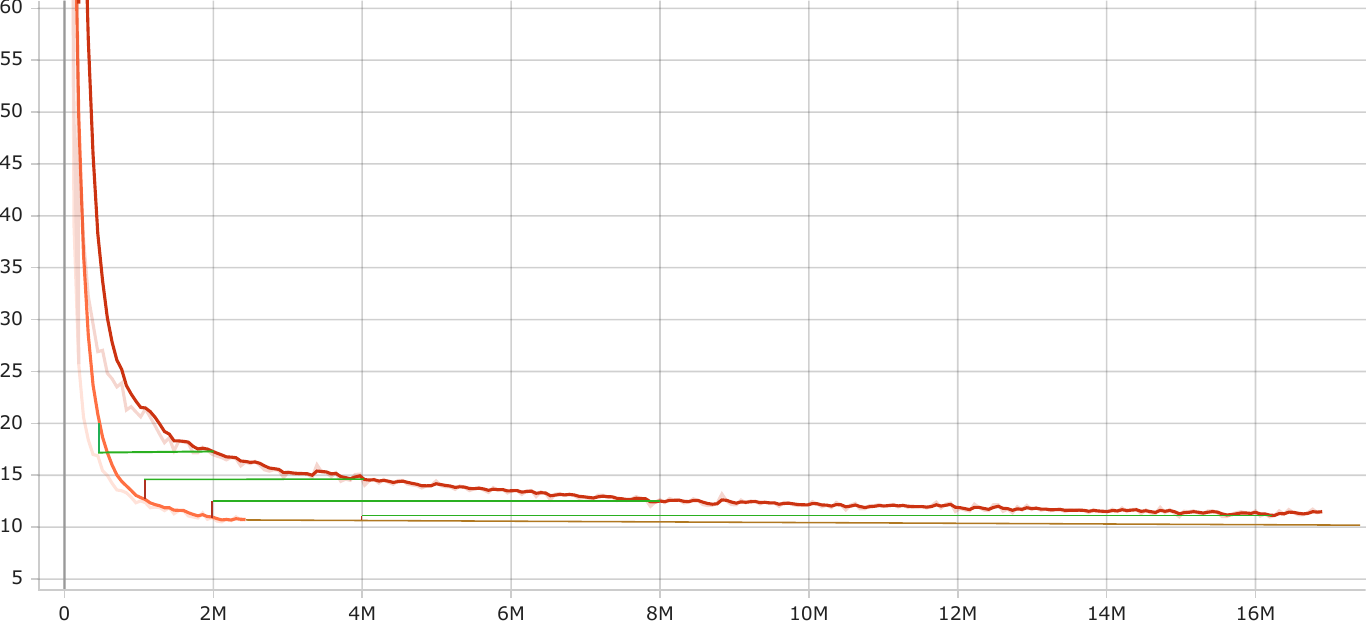}%{images/Curse_of_multilinguality.jpg}   
    \caption{Loss curve on English dev-set. The orange line is related to single-language model, the red was trained with several languages. Green lines show were the models consumed the same amount of English data.}
    \label{fig:curse_of_mult}
\end{figure}

Figure~\ref{fig:curse_of_mult} illustrates the so-called ``curse of multilinguality'': the same amount of English data leads to worse model performance in a multilingual training setup compared to a single-language setup.

\begin{figure}
     \centering     
     \includegraphics[width=0.7\textwidth]{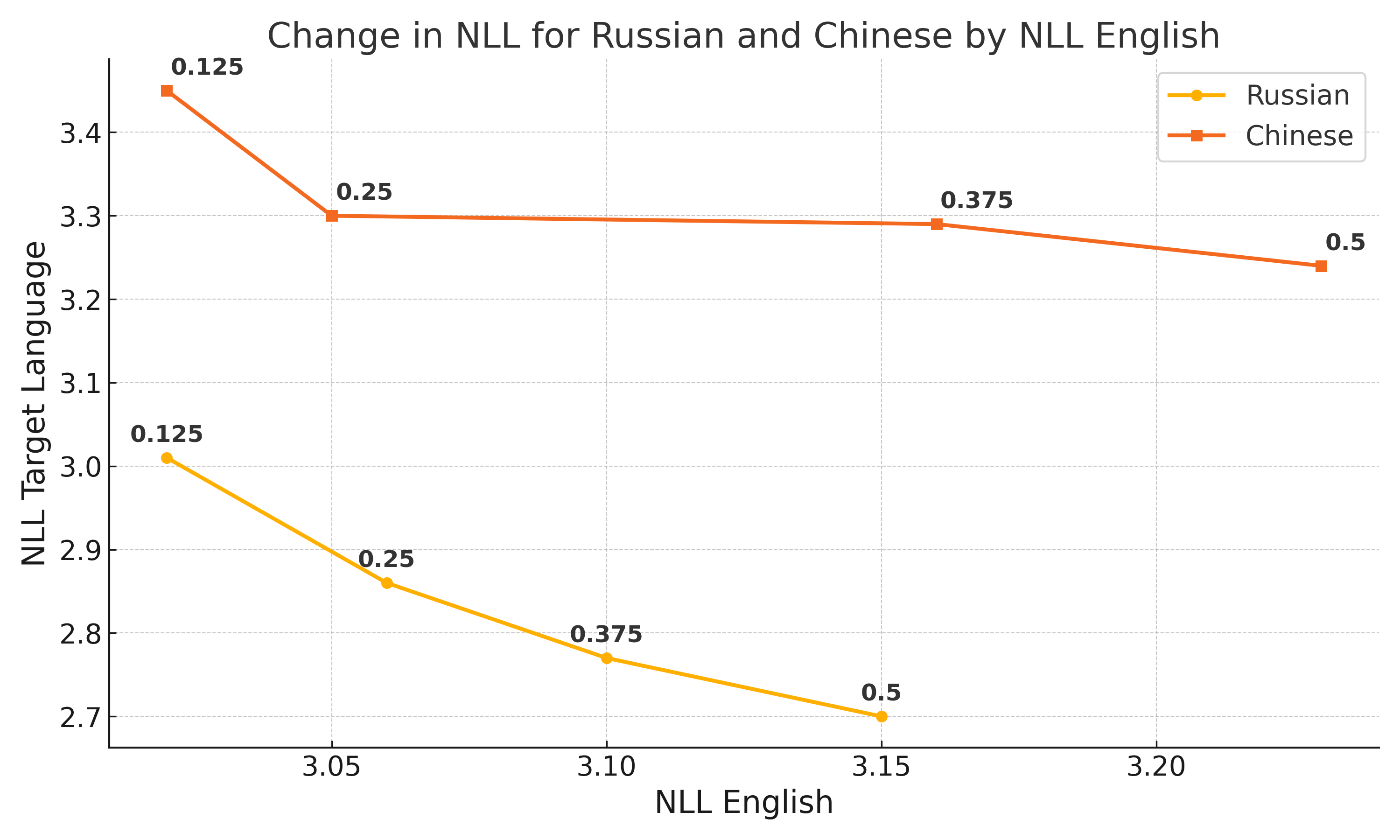}   
    \caption{Final dev loss of the models trained with different proportions of Russian(yellow) and Chinese (orange) with respect to English. We can see that adding more Chinese tokens can have minimal effect to Chinese loss, while the English loss still degrades. For Russian, the Pareto front is more clear.}\label{fig:mixing}
\end{figure}

Figure~\ref{fig:mixing} shows the results of mixing the same amount of English with Russian (yellow) and Chinese (orange) subsamples from the mC4 corpus. The results indicate that adding Chinese leads to less efficient training. This difference can be attributed not only to linguistic similarity, but also to varying data quality within mC4.

We also experimented with changing the distribution of the language during the training based on their size in tokens. The first approach, that we tried, was temperature scaling. Let the empirical distribution be
$p_l = \frac{n_l}{\sum_{l' \in L} n_{l'}},$
where $n_l$ is the number of tokens available for language $l$, then, the temperature scaled distribution is defined as
$q_l = \frac{p_l^{1/t}}{\sum_{l' \in L} p_{l'}^{1/t}},$
where t is the parameter commonly set to 1.43 or 3.33. This method increases the proportion of low-resource languages. If we increase the proportion too much, it may result in overfitting as we do several epochs on the same data. Another approach is UniMax sampling that pre-allocates tokens for low-resource languages based on the allowed max repeats and distributes the remaining budget uniformly across all other languages. We tried both of these approaches, but they led to substantial degradation on high-resource languages. The models were trained on Wikipedia for nearly 50B tokens. In the table, we see those manipulations with the sampling lead to prominent decrease for high-resource languages, but they give some benefits for low-resource and medium-resource languages. As we are more concerned with the high-resource languages, we decided not to up-sample low-resource (Tab.~\ref{tab:sampling_ppl}).

\begin{table}[h]
\caption{Final perplexity on different languages for models trained with various language sampling methods.}
\centering
\resizebox{\linewidth}{!}{
\begin{tabular}{lccccccc}
\toprule
\multirow{2}{*}{\textbf{Sampling Method}} &
\multicolumn{4}{c}{High-resource} & \multicolumn{2}{c}{Mid} & Low \\
\cmidrule(lr){2-5} 
\cmidrule(lr){6-7} 
\cmidrule(l){8-8}
& \textbf{En PPL} & \textbf{Ru PPL} & \textbf{Fr PPL} & \textbf{De PPL} & \textbf{Ar PPL} & \textbf{Bg PPL} & \textbf{Te PPL} \\\midrule
Proportional to number of tokens   & \textbf{7.37} & \textbf{5.68} & \textbf{6.52} & \textbf{7.77} & 6.17 & 6.59 & 3.86 \\
Temperature sampling ($t=1.43$)    & 7.47 & 5.86 & 6.63 & 7.82 & 6.05 & 6.30 & 3.55 \\
Temperature sampling ($t=3.33$)    & 8.42 & 6.01 & 6.78 & 8.34 & \textbf{5.88} & \textbf{5.92} & \textbf{3.22} \\
UniMax (Num repeats = 6)           & 8.68 & 6.14 & 7.05 & 8.72 & 6.15 & 6.68 & 3.87 \\
\bottomrule
\end{tabular}
}
\label{tab:sampling_ppl}
\end{table}

\section{Architecture details}\label{app:architecture}

\begin{table}[H]
\centering
\caption{Architecture details.}
\label{tab:architecture}
\begin{tabular}{lr}
\toprule
\textbf{Parameter} & \textbf{Value}  \\
\midrule
Number format & bf16 \\
Hidden size & 2048 \\
Number of layers & 24 \\
Number of heads & 16 \\
Intermediate size in FFN & 5440  \\
Activation function & SwiGLU \\
Position encodings & ROPE \\
LayerNorm & RMSNorm \\
QK-norm & No \\
Vocab size & 128256 \\
Tied input-output embeddings & Yes\\
Total number of parameters & 1467582464\\
\bottomrule
\end{tabular}
\end{table}

\section{Tokenizers comparison}\label{app:tokenizers}

Table~\ref{tab:tokenizer_comparison} demonstrates the fertility of different tokenizers.

\begin{table}[H]
\centering
\caption{Fertility (number of tokens per word) across languages for different tokenizers.}
\label{tab:tokenizer_comparison}
\resizebox{\linewidth}{!}{
\begin{tabular}{l|r|ccccccccc|c}
\toprule
\textbf{Tokenizer} & \textbf{Vocab size} & \textbf{ru} & \textbf{en} & \textbf{es} & \textbf{de} & \textbf{zh} & \textbf{pl} & \textbf{bg} & \textbf{ja} & \textbf{ar} & \textbf{avg} \\
\midrule
Monolingual        & --      & \textbf{1.21} & \textbf{1.07} & \textbf{1.06} & \textbf{1.32} & 1.50 & --   & --   & --   & --   & --   \\
XGLM               & 256008  & \textbf{1.45} & 1.13 & \textbf{1.17} & 1.49 & \textit{1.28} & \textbf{1.56} & \textbf{1.45} & \textit{1.26} & \textbf{1.20} & \textbf{1.38} \\
GPT-4              & 100277  & 2.76 & \textit{1.07} & 1.38 & 1.73 & 2.08 & 2.19 & 2.88 & 2.06 & 2.82 & 2.16 \\
GPT-4o             & 200019  & \textit{1.60} & \textbf{1.06} & \textbf{1.17} & \textbf{1.42} & 1.39 & 1.88 & 1.87 & 1.50 & \textit{1.22} & 1.52 \\
GPT-3              & 50257   & 6.21 & 1.08 & 1.72 & 2.30 & 3.21 & 3.00 & 5.92 & 2.65 & 4.01 & 3.33 \\
mGPT               & 100000  & 1.65 & 1.29 & \textit{1.21} & 1.47 & \textit{1.28} & \textit{1.77} & \textit{1.81} & \textbf{1.16} & 1.37 & \textit{1.51} \\
Mistral-7B         & 32000   & 2.32 & 1.16 & 1.55 & 1.96 & 1.95 & 2.39 & 2.33 & 2.12 & 3.61 & 2.25 \\
LLaMAv2            & 32000   & 2.12 & 1.21 & 1.48 & 1.82 & 2.42 & 2.23 & 2.22 & 2.29 & 3.67 & 2.25 \\
LLaMAv3            & 128000  & 1.84 & 1.07 & 1.38 & 1.73 & 1.39 & 2.16 & 2.24 & 1.38 & 1.53 & 1.68 \\
Pangu Universal    & 116444  & 1.88 & 1.36 & 1.31 & 1.75 & 1.56 & 1.98 & 1.94 & 1.84 & 1.57 & 1.77 \\
Deepseek-Coder     & 32000   & 2.80 & 1.18 & 1.62 & 2.49 & 1.64 & 3.09 & 2.06 & 2.99 & 5.03 & 2.63 \\
Qwen2              & 151643  & 2.02 & 1.09 & 1.39 & 1.75 & \textbf{1.20} & 2.07 & 2.46 & 1.33 & 1.47 & 1.71 \\
\bottomrule
\end{tabular}
}
\end{table}

\section{MMLU MCF}\label{app:MMLU}
\begin{table}[H]
\centering
\caption{Difference between MMLU MCF performance between base and instructed versions of the models. Training compute (FLOPs) is estimated as $C = 6 \times T \times P$, where $T$ is the number of pre-training tokens and $P$ is the number of model parameters~\citep{hoffmann2022training}. This estimate does not account for the computational resources required for deliberate data generation, which can also be substantial.}
\label{tab:MMLU_MCF}
\vspace{0.5em}
\begin{tabular}{lcccccc}
\toprule
Model family & \textbf{Parameters} & \textbf{Tokens} & \makecell{\textbf{Training}\\\textbf{FLOPs}($\times10^{21}$)} & \textbf{MMLU MCF} & \textbf{Difference} \\
\midrule
Llama3.2-1B-b      & 1.2B & 9T  & 66.96  & 31.2 & -- \\
Llama3.2-1B-it     & 1.2B & 9T  & 66.96  & 45.6 & \textbf{+14.4} \\
Gemma3-1b$^*$      & 1B & 2T  & 12.0  & 23.0 & -- \\
Gemma3-1b-it$^*$   & 1B & 2T  & 12.0  & 39.8 & \textbf{+16.7} \\
Qwen2.5-1.5b       & 1.5B & 18T & 166.32 & 60.0 & -- \\
Qwen2.5-1.5b-it    & 1.5B & 18T & 166.32 & 59.7 & \textbf{-0.3} \\
SmolLM2-b          & 1.7B & 11T & 113.04 & 50.1 & -- \\
SmolLM2-it         & 1.7B & 11T & 113.04 & 49.2 & \textbf{-0.9} \\
Olmo2-4-25-1b-b    & 1.5B & 4T  & 35.52  & 42.6 & -- \\
Olmo2-4-25-1b-it   & 1.B & 4T  & 35.52  & 41.2 & \textbf{-1.4} \\
\textbf{Gamayun-b-old$^\diamond$}    & 1.5B & 2.6T & 26.52  & 29.0 & -- \\
\textbf{Gamayun-b}    & 1.5B & 2.6T & 26.52  & 43.9 & \textbf{+14.9}\\
\textbf{Gamayun-it}   & 1.5B & 2.6T & 26.52  & 46.4 & \textbf{+17.4} \\
\bottomrule
\end{tabular}
\end{table}
$^*$Note that Gemma3 1B was trained with knowledge distillation (2T tokens) instead of next-token prediction. As a teacher, Gemma3 27B was used, which was trained on 14T tokens.

$^\diamond$Gamayun-b-old model is the checkpoint of pretrained Gamayun model before instructions were added in pretrained dataset.

\clearpage

\section{Post-training}\label{app:post-training}
Table~\ref{tab:SFT_composition} shows the final composition of our SFT dataset.
\begin{table}[h]
\centering
\caption{Composition of the SFT dataset. The total dataset contains 2.32M instruction-response pairs from different data sources.}
\label{tab:SFT_composition}
\begin{tabular}{l r r}
\toprule
\textbf{Dataset source} & \textbf{Number of samples} & \textbf{Language}\\ \midrule
\multicolumn{2}{c}{\textit{New datasets}} \\
\cmidrule{1-3}
Synthetic MMLU dataset & 380k & en \\
\midrule
\multicolumn{2}{c}{\textit{Math data}}\\
\cmidrule{1-3}
NuminaMath-CoT (SmolTalk) & 100k & en \\
MetaMathQA & 395k & en  \\
Simple-arithmetic-problems & 120k & en \\
Kolmogorov-3  & 13.6k & ru \\
\midrule
\multicolumn{2}{c}{\textit{Other}} \\
\cmidrule{1-3}
SlimOrca & 500k & en \\
Self-OSS-Starcoder2-Instruct & 50k & en \\
CodeAlpaca & 20k & en  \\
Smol-Rewrite (SmolTalk) & 56.2k & en  \\
Smol-Constraints (SmolTalk) & 36.2k & en  \\
Smol-Summarization (SmolTalk) & 101k & en  \\
SystemChats2.0 (SmolTalk) & 35.9k & en  \\
Explore-Instruct-Rewriting (SmolTalk) & 32k & en  \\
OpenHermes2.5 (SmolTalk) & 100k & en  \\
RuInstruct (OpenOrca-ru part) & 380k & ru \\
Russian-code & 26.3k & ru \\
\midrule

\textbf{Total} & \textbf{2.32M} \\ \bottomrule
\end{tabular}
\end{table}

\clearpage

\clearpage

\section{Multi-lingual benchmarks}\label{app:MBenchmarks}
Here, we describe benchmarks used for evaluation in Arabic, French, Bulgarian, Spanish, German, Italian, Polish and Portuguese languages.

For measuring results in Arabic language we choose ArabicMMLU~\citep{koto2024arabicmmlu}, multi-task language understanding benchmark for the Arabic language, sourced from school exams across diverse educational levels in different countries spanning North Africa, the Levant, and the Gulf regions; ArabCulture~\citep{sadallah2025commonsense}, a commonsense reasoning dataset in Modern Standard Arabic, covering cultures of 13 countries across the Gulf, Levant, North Africa, and the Nile Valley; Open Arabic LLM Leaderboard~\citep{OALL}, which consists of large number of different evaluation tasks that reflect the characteristics of the Arabic language and culturel; and Belebele in Arabic language with Arabic letters and Arabic parts of XNLI and Xstorycloze.

For French language we use the following parts of FrenchBench, a benchmark for evaluating French language models, introduced in~\citep{faysse2024croissantllm}: 
\texttt{Grammar}, \texttt{Vocab}, \texttt{Reading comprehension}, \texttt{XNLI}, \texttt{Hellaswag}, \texttt{Arc-challenge}.
%\texttt{french\_bench\_grammar}, %\texttt{french\_bench\_vocab}, %\texttt{french\_bench\_reading\_comp}, %\texttt{french\_bench\_xnli}, %\texttt{french\_bench\_hellaswag}, %\texttt{french\_bench\_arc\_challenge}
Additionally, we take French part of Belebele, Xwinograd, and several tasks from Le Leaderboard suit~\cite{openllm-French-leaderboard}: BoolQ-fr, MUSR-fr, IF-Eval-fr.

For Bulgarian language we use Belebele in Bulgarian language with Cyrillic letters, INCLUDE-Bulgarian subset, and translated versions of ARC-challenge and Hellaswag from \cite{eu20_benchmark}.

For Spanish language, we evaluate models on Spanish part of Belebele, and tasks present in La Leaderboard~\cite{grandury-etal-2025-la} including Escola~\citep{bel-etal-2024-escola}, Spanish parts of XNLI and XStorycloze.

For German language the LLM evaluation collections are scarce, hence, we limit the model's comparison to several translated benchmarks: translated versions of MMLU, Arc-challenge and Hellaswag, German part of Belebele. Moreover, we evaluate models on a GermanRAG benchmark~\cite{german_rag_dpr_quad}.

For Italian language, in addition to the translated benchmarks, we evaluate models on ItaBench~\cite{moroni-etal-2024-towards} and Evalita-LLM~\cite{magnini2025evalitallmbenchmarkinglargelanguage}.

For Polish language, we use Polish Open Leaderboard's benchmarks~\cite{open-pl-llm-leaderboard} along with the translated versions of ARC-challenge, Hellaswag, MMLU and Belebele.

 For Portuguese, we also use benchmarks from Portuguese Open Leaderboard~\cite{open-pt-llm-leaderboard}.

\section{Multi-lingual Results}\label{app:MResults}
Tables below provide the detailed performance comparison of Gamayun and other 1-2B instructed models across benchmarks in the considered languages. Gamayun achieves the best or near SOTA results on these languages, except for Thai and Chinese. We primarily evaluated model in a zero-shot scenario using lm-harness-eval and lighteval frameworks for the evaluation. We used the following collections of the benchmarks and their respective forks of lm-harness-eval: Evalita-LLM \cite{magnini2025evalitallmbenchmarkinglargelanguage} and ItaBench \cite{moroni-etal-2024-towards} for Italian, La Leaderboard \cite{grandury-etal-2025-la} for Spanish, Le Leaderboard \cite{openllm-French-leaderboard} for French, Open Polish Leaderboard \cite{open-pl-llm-leaderboard}, Open Portuguese Leaderboard \cite{open-pt-llm-leaderboard}.

Qwen3 was evaluated in a no-thinking mode. 

%%%% MULTI_LINGUAL EVALUATION TABLES

\begin{table}[H]
\centering
\caption{Arabic language results}
\vspace{0.5em}
\setlength{\tabcolsep}{5pt}
\begin{tabular}{l c c c c c c}
\toprule
Model& Gemma3 & Llama3.2 & Gamayun & Qwen2.5 & EuroLM  & Qwen3  \\
%\midrule
%\multicolumn{7}{c}{\textit{Arabic}} \\
\midrule
Arabic MMLU   &  39.8 & 37.2 & \underline{47.0}  & \textbf{47.2} & 26.8  & 46.3  \\
Belebele-arb  & 41.78 & 42.11 & \textbf{66.20}  & \underline{65.78} & 28.33 & 62.78 \\
Arabic culture & 35.0 & 30.0 & \textbf{57.2}  & \underline{54.6}& 35.5  & 40.9 \\
Arabic leaderboard & 43.0  & 38.2 & \textbf{49.8}  & 42.7 & 38.9  & 41.5 \\
Xstorycloze  & 56.3  & 50.5 & \textbf{60.5}   & 55.2 & \underline{56.5}   & 52.4  \\
OpenbookQA-ar & 37.58 & 37.17 & \textbf{53.54} & \underline{48.48} & 20.20 & 44.44 \\
SciQ-ar & 41.41 & 42.11 & \textbf{55.68} & \underline{48.44} & 46.93 & 39.40 \\
ACVA & 67.15 & 56.91 & \underline{67.94} & 66.99 & 60.67 & \textbf{69.15} \\ 
\midrule
Avg. & 45.25 & 41.78 & \textbf{57.23} & \underline{53.67} & 39.22 & 49.61 \\
\bottomrule
\end{tabular}
\end{table}

\begin{table}[H]
\centering
\caption{Bulgarian language results}
\vspace{0.5em}
\setlength{\tabcolsep}{5pt}
\begin{tabular}{l c c c c c c c c}
\toprule
Model& Gemma3 & Llama3.2 & Gamayun & Qwen2.5 & EuroLM  & Qwen3  \\
\midrule
Belebele-bul &  34.00  & 36.80 & \textbf{65.56} & 60.67 & 27.56  & \underline{64.33}\\
%Belebele-bul (lm-harness) &  30.89  & 36.80 & \underline{58.2} & \underline{58.2} & 27.56  & \textbf{61.2}\\
Include-bul & 36.4 & 35.0 & \underline{50.4} & 45.5 & 24.7  & \textbf{52.7}  \\
ARC-C (EU20) & 28.92 & 29.43 & \textbf{33.62} & 26.28 & \underline{31.06}  & 29.69  \\
Hellaswag (EU20) & 37.94 & 33.25 & \textbf{48.92} & 35.75 & \underline{48.41}  & 38.24  \\
EXAMS & 30.82 & 25.88 & \textbf{43.43} & 39.71 & 26.06  & \underline{43.41}  \\
\midrule
Avg. & 33.62 & 32.07 & \textbf{48.38} & 41.69 & 31.56 & \underline{45.67} \\
\bottomrule
\end{tabular}
\end{table}

\begin{table}[H]
\centering
\caption{German language results}
\vspace{0.5em}
\setlength{\tabcolsep}{5pt}
\begin{tabular}{l c c c c c c}
\toprule
Model& Gemma3 & Llama3.2 & Gamayun & Qwen2.5 & EuroLM & Qwen3  \\
\midrule
%\multicolumn{7}{c}{\textit{German languag}} \\
\midrule
ARC-C-de & 32.34 & 29.08 & \textbf{35.67} & 32.51 & 31.57 & \underline{34.47} \\
Hellaswag-de & 47.67 & 42.33 & \textbf{49.28} & \underline{44.15} & 43.35 & 42.9 \\
Belebele-de & 41.78 & 48.89 & \underline{68.44} & \textbf{71.78} & 25.33 & 67.00 \\
MMLU (Global) & 36.04 & 39.27 & 40.58 & \underline{47.00} & 26.10 & \textbf{48.65} \\
GermanRAG & 44.37 & 60.15 & \underline{80.25} & \textbf{83.10} & 36.60 & 56.47 \\
\midrule
Avg. & 40.44 & 43.94 & \underline{55.21} & \textbf{55.71} & 32.59 & 49.90 \\
%Avg. &  & & & & & \\
\bottomrule
\end{tabular}
\end{table}

\begin{table}[H]
\centering
\caption{Spanish language results}
\vspace{0.5em}
\setlength{\tabcolsep}{5pt}
\begin{tabular}{l c c c c c c}
\toprule
Model& Gemma3 & Llama3.2 & Gamayun  & Qwen2.5 & EuroLM  & Qwen3  \\
%\midrule
%\multicolumn{9}{c}{\textit{Spanish}} \\
\midrule
ARC-C-es & 32.48 & 31.03 & \underline{37.69} & \textbf{37.78} & 35.38 & 36.15 \\
Belebele-es  & 35.56 & 51.89 & \underline{67.22}  & \textbf{72.78} & 25.67  &  64.33 \\
COPA-es & 65.8 & 60.4 & \underline{70.2} & 69.6 & \textbf{71.0} & 67.4 \\
Escola & 67.33 & 37.99 & \textbf{70.2} & \underline{69.42} & 69.14 &  66.95 \\
Hellaswag-es & 49.24 & 45.98 & \textbf{55.55} & 52.75 & \underline{53.04} & 49.01 \\
MMLU (Global) & 38.17 & 39.57 & 43.32 & \textbf{49.71} & 25.33 & \underline{48.6} \\
OpenbookQA-es & 37.2 & 32.0 & \textbf{38.0} & 36.8 & \underline{37.4} & 36.6 \\
Xnli & 43.45 &  44.98 & \textbf{49.96} & \underline{48.84} & 48.59 &  42.61 \\
Xstorycloze & 62.81 & 60.36  & \textbf{64.53} & 60.23 & \underline{64.26} & 60.03 \\
\midrule
Avg. & 49.30 & 46.17 & \textbf{55.66} & \underline{55.00} & 48.6 & 53.15 \\
\bottomrule
\end{tabular}
\end{table}

\begin{table}[h]
\centering
\caption{Thai language results}
\vspace{0.5em}
\setlength{\tabcolsep}{5pt}
\begin{tabular}{l c c c c c c c c}
\toprule
Model& Gemma3 & Llama3.2 & Gamayun  & Qwen2.5  & Qwen3  \\
%\midrule
%\multicolumn{7}{c}{\textit{Thai}} \\
\midrule
Belebele-th & 39.22 & 41.33 & 55.33 & \textbf{64.33} & \underline{62.00} \\
m3exams & 29.2 & 29.52 & 31.87 & \textbf{38.51} & \underline{37.45} \\
Thai Exams & 27.84 & \underline{32.52} & 28.7 & \textbf{38.94} & 31.93 \\
Hellaswag & 28.42 & 27.93 & 34.56 & \underline{39.53} & \textbf{39.74} \\
XCOPA & 59.00 & 56.20 & 65.60 & \textbf{74.20} & \underline{68.40} \\
XNLI2 & 57.33 & 59.52 & \underline{67.64} & \textbf{71.61}  & 64.29 \\
\midrule
Avg. & 40.17 & 41.17 & 47.28 & \textbf{54.52} & \underline{50.64} \\
\bottomrule
\end{tabular}
\end{table}

\begin{table}[H]
\centering
\caption{Italian language results}
\vspace{0.5em}
\setlength{\tabcolsep}{5pt}
\begin{tabular}{l c c c c c c}
\toprule
Model& Gemma3 & Llama3.2 & Gamayun & Qwen2.5 & EuroLM  & Qwen3  \\
\midrule
%\multicolumn{7}{c}{\textit{Italian}} \\
%\midrule
ARC-C-it & 34.99 & 29.34 & \underline{35.90} & \textbf{37.47} & 32.08 & 35.33 \\
Belebele-it & 42.56 & 48.00 & \underline{67.22} & \textbf{70.44} & 24.56 & 63.89 \\
Hellaswag-it & 47.43 & 42.34 & \textbf{52.63} & 47.55 & \underline{50.75} & 45.61 \\
MMLU (Global) & 37.19 & 38.64 & 41.87 & \underline{46.71} & 27.40 & \textbf{49.13} \\
Evalita-LLM (agg.) & 36.27 & 35.65 & \underline{41.28} & \textbf{43.33} & 33.96 & 40.24 \\
\midrule
Avg. & 39.69 & 38.79 & \underline{47.78} & \textbf{49.1} & 33.75 & 46.84 \\
\midrule
\multicolumn{7}{c}{\textit{ItaBench translated multiple-choice}} \\
\midrule
ARC-E-it & \underline{53.54} & 44.11 & 52.57 & 47.21 & 50.64 & \textbf{57.92} \\
ARC-C-it & \underline{35.16} & 28.80 & \textbf{35.78} & 32.33 & 34.36 & \textbf{35.78} \\
Hellaswag-it & 48.03 & 42.62 & \textbf{52.93} & 47.82 & \underline{52.20} & 46.47 \\
MMLU (cloze) & 30.32 & 29.48 & \underline{32.61} & 31.1 & \textbf{33.2} & 30.21 \\
MMLU (mc) & 38.52 & 37.81 & 42.24 & \underline{49.06} & 24.37 & \textbf{51.82} \\
Winogrande-it & 60.99 & 59.2 & \textbf{62.18} & \underline{61.93} & 61.16 & \textbf{62.18} \\
TruthfulQA (mc1) & \underline{64.14} & 63.51 & 61.62 & \textbf{65.03} & 63.64 & 63.76 \\
TruthfulQA (mc2) & \underline{47.58} & 47.44 & 41.84 & \textbf{49.07} & 44.76 & 43.84 \\
SciQ-it & \underline{53.81} & 45.58 & \textbf{54.21} & 45.08 & 61.42 & 53.40 \\
SciQ-it with passage & 80.41 & 81.42 & \textbf{85.89} & \underline{85.58} & 85.18 & 79.39 \\
PIQA-it & 63.98 & 60.54 & \underline{64.74} & 60.89 & \textbf{65.27} & 61.7 \\
BoolQ-it & \underline{61.25} & \textbf{61.64} & 58.70 & 61.15 & 58.27 & 57.90 \\
GSM8K-mc-it & 52.46 & 56.94 & \textbf{58.61} & 57.62 & 50.64 & \underline{57.92} \\
\midrule
Avg. & 53.09 & 50.70 & \textbf{54.15} & \underline{53.37} & 53.31 & 53.28 \\
\midrule
\multicolumn{7}{c}{\it ItaBench leaderboard tasks} \\
\midrule
BBH-it & 31.61 & 33.24 & \underline{36.74} & \textbf{37.88} & 29.98 & 36.61 \\
GPQA-it & 26.66 & 26.58 & \textbf{30.11} & 26.91 & 24.73 & \underline{27.59} \\
MUSR-it & 35.45 & 36.77 & 36.77 & \underline{37.96} & \underline{37.96} & \textbf{40.34} \\
IfEval-it (avg.) & \textbf{41.73} & 28.98 & \underline{37.48} & 28.69 & 26.18 & 24.22 \\
MMLU-pro-it & 13.20 & 17.48 & 21.03 & \underline{23.39} & 10.26 & \textbf{26.13} \\
\bottomrule
\end{tabular}
\end{table}

\begin{table}[H]
\centering
\caption{Portuguese language results}
\vspace{0.5em}
\setlength{\tabcolsep}{5pt}
\begin{tabular}{l c c c c c c c c}
\toprule
Model& Gemma3 & Llama3.2 & Gamayun  & Qwen2.5 & EuroLM  & Qwen3  \\
\midrule
\multicolumn{7}{c}{\textit{General benchmarks}}  \\
\midrule
Belebele-pt  & 44.33 & 51.22 & \underline{69.89} & \textbf{70.56} & 23.33 &  68.22  \\
MMLU-pt (Global)  &  37.98  &  39.77  &  42.69  &  \textbf{50.01}  &  23.72  &  \underline{46.46}  \\
ARC-C & 30.6 & 30.43 & \textbf{37.61} & \underline{37.09} & 32.99 & 35.9 \\
Hellaswag & 48.69 & 43.02 & \textbf{53.64} & 51.26 & \underline{52.29} & 47.42 \\
\midrule
Avg. & 40.40 & 41.11  & \underline{50.96} & \textbf{52.25} & 33.08 & 46.47 \\
\midrule
\multicolumn{7}{c}{\textit{Portuguese Open Leaderboard}} \\
\midrule
Assin2 RTE & 80.29 & 55.25 & \textbf{88.74} & \underline{88.45} & 55.16 & 86.63 \\
Assin2 STS & 51.35 & 18.2 & \textbf{71.19} & \underline{70.34} & 5.64 & 55.54 \\
bluex & 33.66 & 25.87 & 40.47 & \textbf{51.88} & 21.97 & \underline{50.49} \\
ENEM & 36.46 & 26.73 & 50.66 & \textbf{58.71} & 20.92 & \underline{53.81} \\
faquad NLI & 54.43 & \underline{58.48} & 53.52 & \textbf{67.8} & 21.21 & 43.97 \\
HateBR offensive & \underline{81.14} & 33.65 & \textbf{82.15} & \underline{81.14} & 74.57 & 58.52 \\
OAB Exams & 30.84 & 29.02 & \underline{34.53} & \textbf{43.14} & 23.66 & 32.98 \\
PT Hate Speech & 58.51 & 41.15 & \underline{66.20} & \textbf{68.25} & 52.41 & 50.89 \\
TweetSentBr & 56.71 & 47.78 & \textbf{62.72} & 58.30 & 31.45 & \underline{61.39} \\
\midrule
Avg. & 53.71 & 37.01 & \underline{61.13} & \textbf{64.55} & 34.11 & 54.91
\\
\bottomrule
\end{tabular}
\end{table}

\begin{table}[H]
\centering
\caption{French language results}
\vspace{0.5em}
\setlength{\tabcolsep}{5pt}
\begin{tabular}{l c c c c c c}
\toprule
Model& Gemma3 & Llama3.2 & Gamayun & Qwen2.5 & EuroLM  & Qwen3 \\
%\midrule
%\multicolumn{7}{c}{\textit{French}} \\
\midrule
ARC-C & 34.73 & 29.17 & \textbf{37.47} & \textbf{37.47} & 34.22 &  \underline{35.76}   \\
Belebele-fr & 46.0 & 52.44 & 68.78 & \textbf{74.2} & 25.9 &  \underline{70.3} \\
Grammar & 70.6 & 66.4 & 75.6 & \underline{76.5} & \textbf{77.3} & 68.9 \\
Hellaswag & 48.83 & 43.00 & \textbf{54.36} & 51.15 & \underline{52.85} & 48.35   \\
Vocab & 68.1 & 64.7 & \underline{74.8} & 74.00 & \textbf{75.6} & 70.6   \\
Reading comp. & 59.2 & 56.3 & \textbf{62.0} & \underline{60.6} & \textbf{62.0} & 56.3   \\
XNLI & 33.3 & 33.3 & \underline{57.9} & \textbf{59.1} & 33.4 &  52.7  \\
XWinograd & 60.2 & 65.1 & \underline{67.5} & 61.5 & \textbf{69.9} & \textbf{69.9}  \\
MMLU-fr (Global) & 37.64 & 39.51 & 42.29 & \textbf{49.95} & 27.51 & \underline{48.17} \\
BoolQ-fr & \textbf{80.9} & 67.42 & \underline{79.21} & 77.53 & 68.54 & 71.35 \\
MUSR-fr (avg) & 28.73 & 31.94 & \textbf{33.79} & 30.99 & \underline{33.13} & 28.86 \\
IF-Eval-fr (avg) & 13.47 & \underline{13.62} & \textbf{15.04} & 13.13 & 11.76 & 10.31 \\
\midrule
Avg. & 48.48 & 46.91 & \underline{55.23} & \textbf{55.51} & 47.49 & 52.63 \\
\bottomrule
\end{tabular}
\end{table}

\begin{table}[H]
\centering
\caption{Chinese language results}
\vspace{0.5em}
\setlength{\tabcolsep}{5pt}
\begin{tabular}{l c c c c c}
\toprule
Model& Gemma3 & Llama3.2 & Gamayun & Qwen2.5 & Qwen3  \\
%\midrule
%\multicolumn{7}{c}{\textit{Chinese}} \\
\midrule
Agieval & 28.05 & 26.32 & 33.34 & \textbf{48.13} & \underline{42.14} \\
CEVAL & 34.83 & 41.74 & 44.81 & \textbf{63.03} & \underline{59.16}  \\
CMNLI & 57.14 & 62.55 & 68.9 & \textbf{79.07} & \underline{71.43} \\
C3 & 57.31 & 67.85 & 75.29 & \textbf{84.70} & \underline{81.00} \\
\midrule
Avg. & 44.33 & 49.62 & 55.59 & \textbf{68.73} & \underline{63.43} \\
\bottomrule
\end{tabular}
\end{table}

\begin{table}[H]
\centering
\caption{Polish language results}
\vspace{0.5em}
\setlength{\tabcolsep}{5pt}
\begin{tabular}{l c c c c c c c c}
\toprule
Model& Gemma3 & Llama3.2 & Gamayun  & Qwen2.5 & EuroLM  & Qwen3  \\
%\midrule
%\multicolumn{7}{c}{\textit{Polish}} \\
\midrule
Belebele-pl  &  39.11  &  41.89  &  \textbf{64.89}  &  \underline{63.33}  &  23.56  &  60.89  \\
ARC-C (EU20)   &  30.20 & 28.16  &  \underline{34.04}  &  31.83  &  \textbf{34.56}  &  30.89 \\
Hellaswag (EU20)  &   42.1  &  35.67  &  \underline{46.67}  &  39.01  &  \textbf{48.83}  &  39.5 \\
MMLU-pl (Global)  &  34.75  &  34.8  &  40.47  &  \underline{42.00} &  26.25  &  \textbf{43.47} \\
\midrule
Avg. &  36.54  & 35.13  &  \textbf{46.51}  &  42.59 &  33.3  &  \underline{43.69}  \\
\bottomrule
\end{tabular}
\end{table}

\begin{table}[H]
\centering
\caption{Polish open leaderboard multiple-choice 0-shot with chat template}
\vspace{0.5em}
\setlength{\tabcolsep}{5pt}
\begin{tabular}{l c c c c c c c c}
\toprule
Model& Gemma3 & Llama3.2 & Gamayun  & Qwen2.5 & EuroLM  & Qwen3  \\
%\midrule
%\multicolumn{7}{c}{\textit{Polish}} \\
\midrule
Polemo2-in (acc)  &  67.31  &  46.81  &  \underline{67.87}  &  35.60  &  45.98  &  \textbf{70.36}  \\
Polemo2-out (acc)  &  58.31  &  34.82  &  \textbf{63.16}  &  41.90  &  38.06  &  \underline{59.92}  \\
8tags (acc)  &  54.30  &  17.82  &  \underline{59.68}  &  46.25  &  42.61  &   \textbf{63.16} \\
Belebele (acc)  &  37.44  &  36.33  &  \textbf{62.44}  &  \underline{60.67}  &  23.33  &  56.67  \\
CBD (f1)  &  13.47  &  \underline{14.93}  &  \textbf{15.03}  &  3.33  &  2.69  &  13.96  \\
DYK (f1)  &  \underline{34.74}  &  29.23  &  \textbf{43.97}  &  33.73  &  29.00  &  30.97  \\
Klej-NER (acc)  &  27.41  &  24.25  &  \textbf{36.35}  &  \underline{29.49}  &  6.41  &  8.75  \\
PolQA-rerank (acc)  &  62.83  &  52.66  &  \textbf{68.01}  &  \underline{67.09}  &  51.73  &   50.04 \\
PPC (acc)  &  39.20  &  \underline{39.60}  &  \textbf{47.90}  &  41.40  &  13.50  & 29.20  \\
PSC (f1)  &  \textbf{72.65}  &  46.66  &  \underline{67.42}  &  43.90  &  46.66  &  48.60  \\
\midrule
Avg. &  \underline{46.76}  & 34.31  &  \textbf{53.18}  &  40.33 &  33.3  &  43.16  \\
\bottomrule
\end{tabular}
\end{table}

%%%% END OF MULTI_LINGUAL EVALUATIION TABLES

\newpage
\section{MERA Results}\label{app:MERAResults}
\begin{table}[H]
\centering
\caption{MERA Results}
\label{tab:MERAResults}
\vspace{0.5em}
\setlength{\tabcolsep}{2pt}
\resizebox{1.0\linewidth}{!}{
\begin{tabular}{l | c | c | c | c | c | c | c | c}
\toprule
Model& \shortstack{Gamayun\\1.5B} & \shortstack{Qwen2.5\\1.5B} & \shortstack{ruadapt\\(Qwen2.5)}  & \shortstack{Qwen3\\1.7B} & \shortstack{QVikhr\\(Qwen3)} & \shortstack{meno-tiny-0.1\\(Qwen3)} & \shortstack{Llama3.2\\1.3B}  & \shortstack{VikhrLlama\\(Llama3.2)}  \\
\midrule
RWSD         & 55.4  & 53.8 & 55.0 & 56.2 & 55.0 & 52.7  & 48.5 & 47.3  \\
PARus        & 70.8  & 54.6 & 57.0 & 52.6 & 53.8 & 53.6   & 49.2  & 52.0  \\
RCB          & 50.9/48.1 & 36.8/26.9 & 39.5/34.8 & 48.6/45.7 & 49.8/46.0 & 38.8/31.1 & 37.4/34.4 & 34.7/34.2 \\
MultiQ       & 27.5/12.2 & 27.5/16.3 & 11.9/3.6  & 13.5/5.4  & 18.4/8.1  & 23.4/16.8 & 22.2/11.8 & 21.1/8.8 \\
ruWorldTree  & 70.1/70.0 & 73.7/73.7 & 74.1/74.1 & 81.0/81.0 & 79.6/79.6 & 69.5/69.5  & 42.1/41.9  & 49.1/49.2 \\
ruOpenBookQA & 68.5/68.5 & 66.8/66.8 & 65.8/65.8 & 62.0/62.2 & 63.8/64.0 & 63.0/63.0 & 40.8/40.9 &  41.8/42.0 \\
CheGeKa      & 5.9/3.4   & 2.3/1.2   & 2.2/1.2   & 1.8/0.5   & 2.8/1.2  & 2.1/1.0 & 1.2/0 & 1.5/0.0\\
ruTiE        & 51.6 & 56.0 & 55.2 & 50.9 & 51.9  & 54.9 & 50.0 & 50.6\\
USE          & 7.7  & 6.2  & 6.4  & 11.5 & 12.1  & 7.7  & 3.4 & 4.7\\
MathLogicQA  & 33.8 & 39.4 & 38.1 & 38.6 & 38.5  & 40.4  & 26.8 & 27.6 \\
ruMultiAr    & 14.6 & 22.1 & 19.0 & 19.0 & 19.5  & 24.6 & 16.9 & 15.5\\
LCS          & 11.1 & 6.8  & 5.2  & 10.6 & 9.4   & 6.0 &  9.2 & 10.2\\
ruModAr      & 40.9 & 45.3 & 31.2 & 41.7 & 44.4  & 45.6  & 40.5 & 39.8\\
MaMuRAMu     & 51.7 & 54.4 & 51.9 & 54.3 & 55.3  & 52.3  & 37.0  & 37.7 \\
ruCodeEval   & 9.9/23.0/30.5 & 2.0/1.1/1.8 & 0/0/0 & 2.7/10.8/16.5  & 1.6/6.4/9.8  & 1.0/3.0/6.0  & 14.9/31.7/39.0 & 13.0/26.2/31.7 \\
\midrule
Total & \textbf{38.1} & 35.7  & 33.7  & 36.4 & \underline{36.8} & 35.2  & 29.8 & 30.0 \\
\bottomrule
\end{tabular}}
\end{table}
\newpage
\section{SFT vs DPO results}\label{app:DPOResults}
Table~\ref{tab:DPOresults} provides the comparison between SFT and DPO (Final instructed) version of Gamayun models. We see that DPO improves results across most of benchmarks. 
\begin{table}[h]
\centering
\caption{Results}
\label{tab:DPOresults}
\vspace{0.5em}
\begin{tabular}{l c c}
\toprule
Model & Gamayun SFT & Gamayun DPO   \\
\midrule
MMLU      & 46.2 & \textbf{46.4} \\
MMLU-pro  & 23.8 & \textbf{24.9} \\
Ifeval    & 40.7 & \textbf{56.2} \\
Hellaswag & 47.6 & \textbf{49.3} \\
ARC-E     & 73.6 & \textbf{73.7} \\
ARC-C     & 41.4 & \textbf{42.5}  \\
\midrule
\multicolumn{3}{c}{\textit{Russian}} \\
\midrule
MMLU-ru  &\textbf{41.2} & \textbf{41.2}  \\
Belebele-ru & \textbf{64.0} & 63.1 \\
Xnli-ru  & \textbf{48.4} & 47.8\\
Xstorycloze &  67.8 & \textbf{68.3}\\
Xwinograd  & 65.7 & \textbf{66.7} \\
MERA & 37.1 & \textbf{38.1} \\
\midrule
\multicolumn{3}{c}{\textit{Math}} \\
\midrule
GSM8K & 54.9 & \textbf{56.4}\\
\bottomrule
\end{tabular}
\end{table}

\clearpage

\section{FineWeb-Edu prompts for Russian}\label{app:prompts}
Below we provide the prompts used to label Russian data for training the quality classifier. We used two prompts: one is a direct Russian adaptation of the original prompt, and the other was designed to better capture issues commonly observed in large-scale datasets.
In particular, second prompt was specifically constructed for the classifier to be biased towards long, verbose, self-contained and structured texts and to be biased against advertisements, general filler-like web-pages without much information, and without discriminating scientific literature over fiction or vise versa. Both prompts were additionally tuned to increase the quality of LLM's evaluation and to simplify labels extraction (just searching for ``\textcyr{Оценка:}'' and parsing the following digit allowed for 99.99\% of labeled texts to be parsed correctly).

% \begin{verbatim}
\begin{tcolorbox}[title=Adapted FineWeb-Edu educational score prompt]
\textcyr{Ниже приведен фрагмент веб-страницы. Оцени, имеет ли страница высокую образовательную ценность и может ли быть полезна в образовательном контексте для обучения начиная с начальной до средней школы используя аддитивную оценку от 0 до 5 по системе, описанной ниже. \\
Баллы накапливаются основываясь на удовлетворении каждого из критериев:\\
\\
Добавь 1 балл, если текст предоставляет некоторую базовую информацию, которая относится к образовательным темам, даже если она включает в себя некоторую нерелевантное или неакадемическое содержимое такое как рекламу или промо материал.\\
Добавь еще один балл, если текст рассматривает некоторые элементы относящиеся к образованию но не согласуются с образовательными стандартами. В тексте образовательное содержимое может быть смешано с необразовательным материалом, предлагая поверхностный обзор потенциально полезных тем, или презентуя информацию в неорганизованной манере и бессвязном стиле написания.\\
Присуди третий балл, если текст уместен для использования в образовательных целях и вводит ключевые концепты которые релевантны к школьной программе. Текст связный хотя может не быть всеобъемлющим или может включать в себя некоторую постороннюю информацию. Он может походить на вводную главу учебника или базовое руководство которое подходит для обучения но имеет значимые ограничения такие как обработка концептов которые слишком сложные для учащихся средних классов.\\
Выдай четвертый балл, если текст крайне релевантный и благодатный для образовательных целей для уровня не выше средней школы, обладающие чистый и последовательный стиль написания. Он может быть похож на главу из учебника или руководства, предлагающего существенное образовательное содержимое, включая в себя упражнения и решения с минимальной нерелевантной информацией и концепты не слишком продвинутые для учеников средней школы. Содержимое связно, сконцентрированно и ценно для структурированного обучения.\\
Надели пятый балл, если текст выдающийся в своей образовательной ценности, идеально подходящий для обучения и в начальной, и в средней школе. Текст следует подробным рассуждениям, за стилем написания легко следовать и он предлагает глубокие и тщательные представления о проблематике, лишенный какого-либо необразовательного или запутанного содержимого.\\
\\
Рассуждай и выписывай свои доводы перед тем, как выдать оценку.\\
\\
Текст: '\{\}'\\
Далее:\\
- Кратко обоснуй свою оценку. Формат: "Краткое обоснование: <...>.". Не пиши больше 100 слов.\\
- Напиши оценку. Формат: "Оценка: <число> баллов.". Оценка должна быть выдана ТОЛЬКО в соответствии с критериями. Число должно быть 0, 1, 2, 3, 4 или 5. Не пиши оценку в других форматах.
}
\end{tcolorbox}
% \end{verbatim}

\begin{tcolorbox}[title=General text quality prompt]
\textcyr{Далее представлен текст интернет страницы. Проанализируй и дай оценку от 0 до 5 баллов относительно того, является ли текст информативным и понятным. Баллы накапливаются посредством удовлетворения следующих критериев:\\
- Добавь 0 баллов, если текст не имеет единого смысла.\\
- Добавь 1 балл, если текст предоставляет краткую базовую информацию, без подробностей. Примеры: реклама, описание видео, список товаров, набор названий или отправления вокзала.\\
- Добавь 2 балла, если текст является коротким описанием объекта без ценного мнения, рекламой или набором отзывов в неформальном стиле. Текст может быть информативным, но коротким, обрезанным или недостаточно подробным. Примеры: новостной заголовок, короткий пост в соцсети, описание товара в магазине, набор комментариев к посту.\\
- Добавь 3 балла, если текст правильно структурирован и причинно-следственные связи доведены до конца. Текст должен содержать не меньше 10 предложений. Примеры: новостная статья, выдержка из статьи или закона, неформальный диалог, небольшой блог-пост, сочинение старшеклассника.\\
- Добавь 4 балла, если текст рассматривает тематику со всех сторон, очень подробно погружается в тему от начала до конца или содержит подробные повествования с описаниями. Текст обязан представлять оригинальный взгляд, интересную историю или крайне подробную и доступную информацию. Текст должен содержать не меньше 20 предложений. Примеры: подробная научно-популярная статья, небольшой рассказ, информативный блог.\\
- Добавь 5 баллов, если текст идеально структурирован, детализирован, информативен, стилистически выдержан, имеет эталонную грамматику и синтаксис и не имеет ошибок любого вида. Примеры: обширный монолог по некой теме, отрывок классического произведения литературы, научной статьи.\\
\\
Рассуждай и выписывай свои доводы перед тем, как выдать оценку. Подробно проверь текст на соответствие критериям.\\
\\
Текст: '\{\}'\\
Далее:\\
- Посчитай количество предложений в тексте в формате: "Длина: <длина> предложений.".\\
- Кратко обоснуй свою оценку. Формат: "Краткое обоснование: <...>.". Не пиши больше 100 слов.\\
- Напиши оценку. Формат: "Оценка: <число> баллов.". Оценка должна быть выдана ТОЛЬКО в соответствии с критериями. Число должно быть 0, 1, 2, 3, 4 или 5. Не пиши оценку в других форматах.
}
\end{tcolorbox}

\section{Examples of generations of Gamayun in comparison to Qwen2.5-1.5B}\label{sec:examples}

Below are example generations from Qwen (former) and Gamayun (latter), with errors marked in color:
\textcolor{hallucination}{\textbf{red}} = hallucination, 
\textcolor{grammar}{\textbf{blue}} = grammar error.
%,\textcolor{logic}{\textbf{orange}} = logical/mathematical error.

Both models show a tendency to hallucinate(Example~\hyperref[ex1:hamlet]{1}), although their short answers are sometimes correct (e.g., Examples~\hyperref[ex4:planets]{4}, \hyperref[ex5:jupyter]{5}). Compared to Gamayun, Qwen more frequently struggles with complex Russian morphology, producing non-existent or inappropriate word forms (e.g., Example~\hyperref[ex2:anna]{2}). Both models successfully solved the simple mathematical task (Example~\hyperref[ex3:math]{3}).

% ===== PAIR 1: Hamlet =====
\subsection*{Example 1: Who is Hamlet?}\label{ex1:hamlet}

\noindent\textbf{Qwen:}
\begin{quote}
Hamlet is a fictional character from William Shakespeare's play \emph{Hamlet.} The play was written around 1602 and is one of Shakespeare's most famous works.

Prince of Denmark: \textcolor{hallucination}{Hamlet inherits the throne from his uncle Claudius}, who has murdered his father to become king. 

Revenge plot: He seeks revenge against those responsible for his father's murder (\textcolor{hallucination}{Clodius}) and uncle Claudius.  

Death of Polonius: One of the play's most famous scenes occurs when \textcolor{hallucination}{he confronts Laertes about his brother Polonius' death by poison}. 

Prologue: \textcolor{hallucination}{The prologue introduces Hamlet as a character who will cause "The grave to be glad" and "the damned to weep"}.
\end{quote}

\noindent\textbf{Gamayun:}
\begin{quote}
Hamlet is a character from William Shakespeare's tragedy "Hamlet, Prince of Denmark." He is the son of King Hamlet and Queen Gertrude, and the nephew of King Claudius. Hamlet is often seen as a complex and tragic figure, dealing with a series of revenge fantasies and personal vendettas.

Here’s a brief overview of his character:

1.  \textbf{Role in the Play}: Hamlet is the main protagonist of the play, which is a retelling of the story of Hamlet, Prince of Denmark, a prince who is $\color{red}{\text{murdered by his uncle}}$, Claudius, who then marries his widow, Gertrude.

2.  \textbf{Motives and Intentions}: Hamlet is driven by a deep-seated desire for revenge against his uncle Claudius. His revenge is fueled by the belief that Claudius murdered his father, King Hamlet, and then married Gertrude, who is now his widow.

3.  \textbf{Character Development}: Throughout the play, Hamlet undergoes a series of psychological and emotional transformations. He is initially consumed by grief and rage, but as the play progresses, he begins to question his own sanity and the motives behind his actions.

4.  \textbf{Relationships}: Hamlet's relationships with his family and friends are complex. He is torn between loyalty to his father and $\color{red}{\text{uncle}}$ and his desire for vengeance. His relationships with his mother, Gertrude, and his $\color{red}{\text{sister, Ophelia}}$, are particularly poignant.

5.  \textbf{Themes}: The play explores themes of power, deception, madness, and the human condition. Hamlet's struggle to uncover the truth about his father's death and seek justice is a central theme.

Hamlet's character is a multifaceted and deeply studied figure in Western literature, known for its exploration of human emotions, motivations, and the limits of human understanding.
\end{quote}

\subsection*{Example 2: \textcyr{Кто такая Анна Каренина?}}\label{ex2:anna}

\noindent\textbf{Qwen:}
\begin{quote}
\textcyr{
Анна Каренина — \textcolor{blue}{литературная персонажница} романа Льва Толстого \textcolor{red}{``Война и мир'}. Вот основные характеристики этого персонажа:

1. Родилась в \textcolor{red}{1859} году.

2. Вступила в брак с \textcolor{red}{Петром} Карениным, но была его \textcolor{blue}{первым женою}.

3. \textcolor{red}{Позже} \textcolor{blue}{ушла замуж} \textcolor{red}{за Александра Долгорукова}.

4. \textcolor{red}{Умерла при беременности от} \textcolor{blue}{преждевременного абортания} \textcolor{red}{1901 года}.

5. Она — одна из главных героинь романа, символизирующая духовные проблемы русского общества \textcolor{red}{начала XX века} 

6. Каренина является прототипом для многих последующих литературных персонажей — \textcolor{red}{от Марии Измайловны Волконской до Эллы Булгаковой}.
}
\end{quote}

\noindent\textbf{Gamayun:}
\begin{quote}
\textcyr{Анна Каренина — героиня романа Льва Толстого ``Анна Каренина''. Она мать двоих детей, замужем за богатым мужем. Анна — женщина, глубоко уязвимая и самолюбивая, любящая и привязчивая. Ее история — это сложная драма любви, предательства и моральных дилемм. Анна борется с собственной виной, неуверенностью в правильности своих действий и стремлением к идеалу любви.
}
\end{quote}

\subsection*{Example 3: What is larger: 2 power 3 or 3 power 2?}\label{ex3:math}

\noindent\textbf{Qwen:}
\begin{quote}
To compare \(2^3\) and \(3^2\), we first calculate the values of each expression:

- For \(2^3\):
  \[
  2^3 = 2 \times 2 \times 2 = 8
  \]

- For \(3^2\):
  \[
  3^2 = 3 \times 3 = 9
  \]

Clearly, \(9\) is larger than \(8\). Therefore, \(3^2\) (which is \(9\)) is the larger number.
\end{quote}

\noindent\textbf{Gamayun:}
\begin{quote}
To compare \(2^3\) and \(3^2\), we calculate each power separately:
\begin{enumerate}
    \item \textbf{Calculate \(2^3\):}
   \[
   2^3 = 2 \times 2 \times 2 = 8
   \]
   \item \textbf{Calculate \(3^2\):}
   \[
   3^2 = 3 \times 3 = 9
   \]
\end{enumerate}

Now, comparing the results:
\begin{itemize}
    \item \(2^3 = 8\)
    \item \(3^2 = 9\)
\end{itemize}

Since \(9 > 8\), it follows that \(3^2 > 2^3\).

Thus, \(3^2\) is larger than \(2^3\).

Conclusion:
\[
\boxed{3^2 > 2^3}
\]
\end{quote}

\subsection*{Example 4: Give me the list of planets in Solar system}\label{ex4:planets}

\noindent\textbf{Qwen:}
\begin{quote}
Mercury, Venus, Earth, Mars, Jupiter, Saturn, Uranus, Neptune
\end{quote}

\noindent\textbf{Gamayun:}
\begin{quote}
The planets in our Solar System, in order from the Sun, are:
\begin{enumerate}
    \item Mercury
    \item Venus
    \item Earth
    \item Mars
    \item Jupiter
    \item Saturn
    \item Uranus
    \item Neptune
\end{enumerate}

\textbf{Note:} Pluto was reclassified as a "dwarf planet" by the International Astronomical Union in 2006.
\end{quote}

\subsection*{Example 5: Which planet is the largest?}\label{ex5:jupyter}

\noindent\textbf{Qwen:}
\begin{quote}
Jupiter
\end{quote}

\noindent\textbf{Gamayun:}
\begin{quote}
The largest planet in our solar system is Jupiter. It is a gas giant, which is a type of planet that is primarily composed of hydrogen and helium, and has a thick atmosphere of gases. Jupiter is about 11 times larger in diameter than Earth, and it is also the fifth planet from the Sun. Its mass is about 318 times that of Earth, making it the heaviest planet in our solar system.
\end{quote}

\end{document}